\documentclass[10pt,letterpaper,compsoc,conference]{style/iiswc22}
\usepackage{cite}
\usepackage{amsmath,amssymb,amsfonts}
\usepackage{algorithmic}
\usepackage{graphicx}
\usepackage[dvipsnames]{xcolor}
\usepackage[final]{microtype}
\usepackage[italic]{mathastext}
\usepackage{libertine}
\usepackage[T1]{fontenc}
\usepackage{textcomp}
\usepackage[varqu,varl]{zi4}
\usepackage[all]{nowidow}
\usepackage[auth-lg,affil-it]{authblk}
\usepackage[keeplastbox]{flushend}
\usepackage{formatting/packages}
\usepackage{formatting/names}
\usepackage{formatting/macros}
\usepackage{formatting/results}
\usepackage{booktabs}
\usepackage{relsize}
\makeatletter
\def\footnoterule{\kern-3\p@
  \hrule \@width 2in \kern 2.6\p@} 
\makeatother
\newcommand\blfootnote[1]{%
  \begingroup
  \renewcommand\thefootnote{}\unskip\footnote{#1}%
  \addtocounter{footnote}{-1}%
  \endgroup
}

\begin{document}

\title{\oracle: A Graph Neural Network Model for\\Basic Block Throughput Estimation}
\author[]{%
Ond\v{r}ej S\'{y}kora\quad{}Phitchaya Mangpo Phothilimthana$^\dagger$\quad{}Charith Mendis$^*$\quad{}Amir Yazdanbakhsh$^\dagger$\\
\small{Google Research\quad{}$^\dagger$Google Research, Brain Team\quad{}$^*$University of Illinois at Urbana Champaign}\\\vspace{0.3cm}
\scriptsize
{\texttt{\href{mailto:ondrasej@google.com}{ondrasej@google.com}, \href{mailto:mangpo@google.com}{mangpo@google.com}, \href{mailto:charithm@illinois.edu}{charithm@illinois.edu},
\href{mailto:ayazdan@google.com}{ayazdan@google.com}}
}
}

\maketitle
\begin{abstract}
\label{sec:abstract}
Analytical hardware performance models yield swift estimation of desired hardware performance metrics.
However, developing these analytical models for modern processors with sophisticated microarchitectures is an extremely laborious task and requires a firm understanding of target microarchitecture's internal structure.
In this paper, we introduce {\sc{\oracle}}\footnote{\textbf{\oracle}: A \textbf{GRA}ph \textbf{N}eural network model for bas\textbf{I}c block \textbf{T}hroughput \textbf{E}stimation}, a new machine learning model that estimates the throughput of basic blocks across different microarchitectures.
{\sc{\oracle}} uses a graph representation of basic blocks that captures both structural and data dependencies between instructions.
This representation is processed using a graph neural network that takes advantage of the relational information captured in the graph and learns a~rich neural representation of the basic block that allows more precise throughput estimation.
Our results establish a new state-of-the-art for basic block performance
estimation with an average test error of 6.9$\%$ across a wide range of basic blocks and microarchitectures for the x86-64 target.
Compared to recent work, this reduced the error by 1.7$\%$ wile improving training and inference throughput by approximately 3.0$\times$. 
In addition, we propose the use of multi-task learning with independent multi-layer feed forward decoder networks. Our results show that this technique further improves precision of all learned models while significantly reducing per-microarchitecture training costs.
We perform an extensive set of ablation studies and comparisons with prior work, concluding a set of methods to achieve high accuracy for basic block performance estimation.
\end{abstract}

\section{Introduction}
\label{sec:intro}
A basic block is a sequence of instructions with neither incoming nor outgoing branches. Basic blocks are natural input objects to many code optimization algorithms because the instructions of a basic block can be modified, as long as the invariants at the beginning and at the end of the basic block are preserved.
See Table~\ref{table:bb} for an example basic block.
Accurate and fast performance estimation of basic blocks is often crucial at the various stages of compilation and software optimization~\cite{lozano2012constraint,stephenson2003meta,mcgovern1999scheduling,nuzman2006autovectorization,jia2019optimizing,narayanan2019pipedream,narayanan2020heterogeneity} because real hardware measurements are expensive to collect and tedious to obtain.
For example, various performance estimation methods are used for inlining~\cite{trofin2021mlgo}, register allocation~\cite{lozano2012constraint}, fusing~\cite{absinthe}, hardware-software co-design~\cite{yazdanbakhsh2021apollo,zhou2022towards,hegde2021mind,kumar2021data}, and critical path analysis~\cite{laukemann2019automatic}.
To provide a fast performance estimation, hand-tuned analytical models~\cite{llvm-mca:web:2019,llvm-sim:web:2020,iaca:web:2019,uica}, tailored for one or few sets of microarchitectures, have been developed.
However, these analytical models are often lack generality across different processors and require domain expertise and thorough knowledge of internal organization of microarchitectural components, which are generally obscured by hardware companies.
Even with sufficient domain knowledge, developing a complete and thorough analytical model for modern processors is an error-prone and work-intensive task.
In addition, due to the increasing complexity of modern microarchitectures, these analytical models may overlook some corner cases in performance estimation and underperform in generalizing the estimation to these cases.
As such, using the analytical models can mislead the optimization algorithm and yield sub-optimal solutions.

\niparagraph{Learned models for throughput estimation.}
To address the aforementioned challenges, a handful of work delegated the task of performance estimation to machine learning~\cite{zhou2022towards,hegde2021mind,trofin2021mlgo}.
For basic block throughput estimation specifically, \ithemal~\cite{ithemal:icml:2019} uses a machine learning model based on a sequential Long-Short Term Memory (LSTM) to learn a representation of basic blocks followed by a linear transformation to predict the throughput values.
While \ithemal~\cite{ithemal:icml:2019} delivered a notable accuracy improvement across multiple x86-64 microarchitectures compared to analytical models of the time, it represents a basic block as a sequence of instructions without any additional information about its structure.
We argue that adding information such as data dependency could contribute to the inductive bias of a model and enables the model to reason about code with higher accuracy.

\niparagraph{Graph-based representation learning of basic blocks.}
\blfootnote{© 2022 IEEE.  Personal use of this material is permitted.  Permission from IEEE must be obtained for all other uses, in any current or future media, including reprinting/republishing this material for advertising or promotional purposes, creating new collective works, for resale or redistribution to servers or lists, or reuse of any copyrighted component of this work in other works.}
Data and control flow in basic blocks can be naturally expressed using a graph~\cite{ncf:iclr:2019}.
This paper sets out to use graph neural networks on this representation of code to learn an expressive representation of basic blocks.
The proposed representation learning method, dubbed \oracle, does not commit to any feature engineering of the input basic blocks.
Compared to prior work, we believe that leveraging a graph representation is a more natural and intuitive approach to represent basic blocks, better capturing the dependencies and interactions between instructions.

\oracle outperforms \ithemal~\cite{ithemal:icml:2019} in terms of accuracy and establishes a new state-of-the-arts results on x86-64 basic block throughput estimation.
We evaluate \oracle for the task of throughput estimation, achieving a new state of the art accuracy, with a nearly $1.7\%$ lower MAPE across multiple x86-64 microarchitectures, compared to the most recent prior work~\cite{ithemal:icml:2019}.
We argue that using a graph representation of basic blocks is a key contributing factor in achieving higher prediction accuracy, which is a direct consequence of better generalization to unseen basic blocks.

\niparagraph{Multi-task throughput estimation model.}
While there are differences in performance of different microarchitectures, there are often also many similarities because of how their design evolved, but also due to instruction set semantics that are microarchitecture-independent.
Multi-task learning~\cite{caruana1997multitask} is a technique that uses a collection of related tasks to train the same model. By exploiting the relatedness of the tasks, the model learns a better internal representation of the problem domain and it often leads to improved performance on the individual tasks.
To our best knowledge, existing work \cite{ithemal:icml:2019} did not take full advantage of these similarities and focused on developing or training a~separate model for each target microarchitecture.
We argue that the similarities between microarchitectures can be exploited to achieve faster training and learning richer representations of code.
To this end, we propose a multi-headed task-dedicated representation learning where the graph network is shared by all microarchitectures and each head is trained for a different microarchitecture.

We evaluated a multi-task model against models that were trained only for a single microarchitecture.
Our results demonstrate that it is feasible to learn a shared representation of basic blocks that support performance predictions for all target microarchitectures.
The computational costs of training a model supporting multiple microarchitectures are only marginally higher than the cost of training a single single-task model.
In addition, we found that employing multi-task learning further reduces the prediction errors on all microarchitectures compared to training exclusively on data from a single microarchitecture.
\section{Motivation and Background}
\label{sec:motivation}
\begin{table}[t!]
\caption{An example basic block in x86-64 assembly from the \bhive dataset~\cite{bhive:iiswc:2019}.}
\small
\begin{center}
\begin{tabular}{|lll|}\hline
0: & CMP    & R15D, 1 \\
1: & SBB    & EAX, EAX \\
2: & AND    & EAX, 0x8 \\
3: & TEST   & ECX, ECX \\
4: & MOV    & DWORD PTR[RBP - 3], EAX \\
5: & MOV    & EAX, 1 \\
6: & CMOVG  & EAX, ECX \\
7: & CMP    & EDX, EAX \\
\hline
\end{tabular}
\end{center}
\label{table:bb}
\end{table}
\subsection{Manual Tuning of Simulator Parameters}
Recent work~\cite{uica} proposes an analytical model to predict the throughput of basic blocks for Intel microarchitectures with sufficient accuracy (< 1$\%$).
To develop this analytical model, the authors performed a detailed study of the underlying Intel microarchitectures and \textit{manually} tuned the microarchitecture-specific parameters of their simulator to match the ground-truth values.
The suggested analytical model establishes stronger baselines for learned throughput estimation across a limited set of microarchitectures and provides interpretable insights about the underlying bottlenecks of the target microarchitecture.
However, the hand-tuned analytical models generally suffer from:
\textbf{(1)} a lack of generality across wide-range of unseen microarchitectures,
\textbf{(2)} a tedious task of maintaining such an analytical model after each generation of microarchitectures,
and finally \textbf{(3)} the demand of expert knowledge about the details of the underlying microarchitecture.
On the other hand, learned models (such as our work and \ithemal \cite{ithemal:icml:2019}), marginally trade off prediction accuracy and interpretability of the results for generality across wide range of microarchitectures and eliminating the need for expert knowledge in the development process.
In summary, analytical and learned models have different objectives and could be beneficial in downstream tasks with different objectives.

\subsection{Learned Model for Throughput Estimation}

\ithemal~\cite{ithemal:icml:2019}, the most recent learned model for basic block throughput estimation, formulates the throughput estimation problem as a regression problem with the objective to minimize the mean absolute percentage error between ground-truth data (obtained from hardware measurements) and the output of the learned model.
It employs a two-level LSTM~\cite{lstm} network that generates an embedding vector for each input basic block.
The objective of the first level LSTM network is to generate an embedding vector for each instruction of the input basic block.
The second level uses the instruction embedding vectors to compute an embedding vector for the whole basic block.

In the input, \ithemal receives a sequence of instructions for each basic block (e.g. \qk{SBB EAX, EAX}, as illustrated in Table~\ref{table:bb}). 
When presenting instructions to the model, \ithemal tokenizes each assembly instruction into (1) instruction mnemonic, (2) input operands, and (3) output operands.
For example, \qk{SBB EAX, EBX} is tokenized as \qk{SBB | <S> | EAX | EBX | <D> | EAX | <E>} tokens, where \qk{<S>}, \qk{<D>}, and \qk{<E>} are special tokens that separate the three groups of tokens.
Each token is mapped to a learned embedding vector (each embedding vector is a real-valued vector of a fixed size), and these vectors are fed to the first-level LSTM network.
Finally, the generated instruction embeddings pass through the second LSTM layer to obtain an embedding vector per basic block.
The generated basic block embedding is then passed to a decoder network to obtain an estimation of the basic block throughput. In the \ithemal model, the decoder is a dot product of the basic block embedding vector with a vector of learned weights.

While \ithemal demonstrates a promising path forward for performance estimation of basic blocks, its input data format presents the instructions to the model linearly, as they are laid out in memory, and it relies on the model and the training process to discover dependencies between the instructions on it own.
Since these dependencies are well defined and easy to extract using existing tools, we suggest including them in the basic block representation explicitly, to guide the computation of the model.
This work wields graphs as a natural and intuitive way to represent basic blocks and the underlying dependencies, expecting that a graph neural network model will be able to benefit from the additional information and produce more precise throughput estimates.

\subsection{Graph Neural Network}
\label{sec:background}
The family of graph neural networks (GNNs)~\cite{gori2005new,scarselli2005graph,kipf2016semi,wu2020comprehensive,graphnet:arxiv:2018} has shown to be effective in a diverse range of applications and domains~\cite{park2019estimating,leclair2020improved,gilmer2017neural,bajaj2019g3raphground}.
Generally, GNNs yield promising results in applications with highly structured inputs where the relationships between elements of the input can be easily expressed using a graph.
The main objective of a GNN is to learn to map the information structured as a graph into an embedding space (a vector representation).
In a nutshell, the learning process of a GNN model consists of propagating information between graph nodes and edges via multiple message passing iterations, followed by an aggregation step.
At each message passing iteration, the node and edge embeddings are updated according to received messages from their neighbors in the graph.
The final learned embeddings are then employed in downstream tasks such as regression, classification, and ranking.
\section{\oracle Model Architecture}
\label{sec:model}
The \oracle model is composed of the following building blocks:
\begin{itemize}
    \item \niparagraph{Graph encoding of basic blocks $\rightarrow$}
    The first step in \oracle is to transform basic blocks into a graph representation and convert the instructions and basic block dependencies into node and edge labels according to their types.
    The constructed graph representation for basic blocks is used as input to the graph neural network.
    \item \niparagraph{Graph neural network $\rightarrow$}
    Next, \oracle uses a GNN model with the objective to learn an expressive embedding for each basic block.
    As part of the training process, the GNN model iteratively exchanges relevant information between basic block elements with the objective of computing the embedding vectors.
    \item \niparagraph{Decoder network $\rightarrow$}
    Each instruction embedding vector passes through an additional decoder network with non-linearity that computes the contribution of the instruction to the basic block throughput.
    \oracle predicts the final throughput values for each basic block by adding all individual instructions' contributions to the overall throughput.
    \item \niparagraph{Multi-task decoder network $\rightarrow$}
    The multi-task version of \oracle uses a multi-task decoder that predicts the throughput values across multiple microarchitectures simultaneously.
    Other parts of the model are shared across all target microarchitectures.
    Intuitively, the task of the shared parts is to learn an internal representation of basic block structure, while the decoder networks are responsible for throughput estimation.
\end{itemize}

\subsection{Graph Encoding of Basic Blocks}
\label{sec:graph-encoding}
\begin{figure}[t]
    \centering
    \begin{verbatim}
    MOV RAX, 12345
    ADD DWORD PTR [RAX + 16], EBX
    \end{verbatim}
    \includegraphics[width=0.45\textwidth]{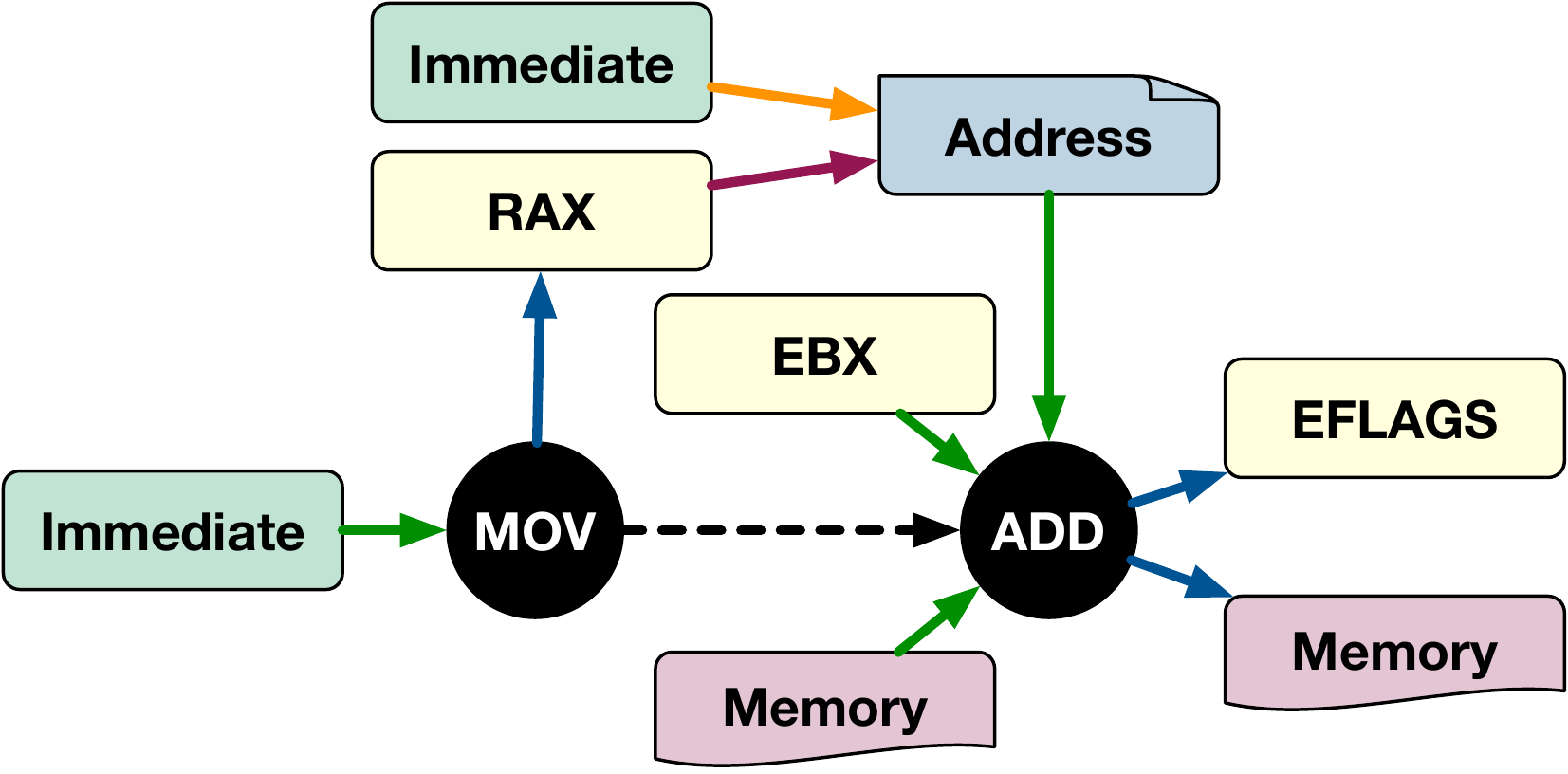}
    \caption{An example basic block with two instructions and its graph representation. The first instruction stores an immediate value (12345) to a register (RAX). The second instruction adds a 32-bit value from the register EBX to a 32-bit value in memory at the address RAX + 16.}
    \label{fig:model:gnn-example}
\end{figure}

We model each basic block as a dependency graph inspired by \cite{ncf:iclr:2019}, but using a more compact format.
The \oracle graph is designed to capture the semantic relationships between instructions as well as the type and category of instructions and registers.
The nodes of the graph consist of a set of instruction and value nodes (e.g. values in registers, immediate values, etc.), whereas the edges indicate data and structural dependencies between the instructions and values represented by the nodes.
Figure~\ref{fig:model:gnn-example} shows an example basic block in the \oracle encoding.

Each node of the graph corresponds to an element of the assembly language, similar to one token in the \ithemal model~\cite{ithemal:icml:2019}. Broadly, we can categorize node types into two groups: \emph{instruction nodes} that represent instructions, and \emph{value nodes} that model the input and output values passed between instructions.
Table~\ref{table:node_types} summarizes the node types in \oracle graph representation and assembly language tokens that can be associated with them.
We represent each assembly instruction by a unique \emph{instruction mnemonic} node.
Infrequently, an assembly instruction may have \emph{prefixes} that modify their behavior, such as \qk{LOCK} or \qk{REP}.
We represent each prefix by a separate graph node that is connected to the instruction mnemonic node by an edge. 

Each instruction node is connected to zero or more \emph{value nodes} representing the instruction operands. The operands are values stored in registers or memory, immediate values, and results of address computation.
Each value node has zero or one incoming edge from the instruction mnemonic node of the instruction that produces it (no incoming edge means that the value is not produced by an instruction of the block), and zero or more outgoing edges to instructions that consume the value.
These edges represent the data dependencies between instructions.
The token associated with a value node is the name of a register if the value is stored in a register, or a special token if the value is stored in memory, it is an immediate value, or it is the result of an address computation.

Note that the nodes represent \emph{a value} in a storage location, not the storage location itself and the graph may contain multiple value nodes with a given register name, if multiple instructions in the block write to this register.
For example, in Figure~\ref{fig:model:gnn-example}, register \qk{RAX} is a destination operand for \qk{MOV} instruction and is used as a source operand to calculate the memory address for \qk{ADD} instruction.
In the same example, you can see two different \qk{Memory} nodes; one is used as an input operand, the other as an output operand. Since the value written by the \qk{ADD} instruction maybe different from the value it reads, they are represented as two distinct nodes.
Table~\ref{table:edge_types} summarizes the list of all edge types in the \oracle graph representation. 
In a nutshell, the existence of an edge between two graph nodes captures the semantic relationships of the connected nodes as well as their sequential ordering. As such, all the edges in the graph are directed.
The last column in Table~\ref{table:edge_types} depicts the color code that we used to show the dependency between graph nodes in the example of a basic block in Figure~\ref{fig:model:gnn-example}.

\subsection{Graph Neural Network}
The objective of a graph neural network model is to learn representative feature vectors for graph nodes and edges that express their underlying characteristics in a latent space.
The graph neural network propagates information between graph elements through message passing iterations.
Before training starts, the graph elements (e.g. nodes and edges) are initialized to unique vector values that are representative of a particular property of each graph element.
\oracle uses the “full GN block” architecture as described in Section 4.2 of \cite{graphnet:arxiv:2018}.
In each message passing iteration, all feature vectors are updated using Algorithm~1 from \cite{graphnet:arxiv:2018}, employing multi-layer feed forward ReLU networks with residual connections~\cite{residual-learning:cvpr} and layer normalization~\cite{ba2016layer} at input as update functions.
The initial values of the feature vectors of elements of the graph depend on the type of the element and the associated elements of the assembly language:
\begin{table}[t!]
\caption{The node types in \oracle graph representation.}
\small
\begin{center}
\small
\begin{tabular}{p{2.8cm}p{4.5cm}}
\hline
\textbf{Node Type}      & \textbf{Token} \\
\hline
\underline{\textbf{Instruction Nodes}} & \\
\tabindent Mnemonic             & The mnemonic of the instruction (e.g. \texttt{ADD}). \\
\tabindent Prefix               & The prefix of an instruction (e.g. \texttt{LOCK}) \\\midrule
\underline{\textbf{Value Nodes}} \\
\tabindent Register                & Register name (e.g. \texttt{RBX}). \\
\tabindent FP immediate value      & Special token shared by all floating-point immediate values. \\
\tabindent Immediate value         & Special token shared by all immediate value nodes. \\
\tabindent Address computation     & Special token shared by all address computation nodes. \\
\tabindent Memory value            & Special token shared by all values stored in memory. \\
\hline
\end{tabular}
\end{center}
\label{table:node_types}
\end{table}
\begin{table}[t!]
\small
\renewcommand{\arraystretch}{1.1}
\centering
\caption{The edge types in \oracle graph representation.}
\label{table:edge_types}
\resizebox{0.49\textwidth}{!}{
\begin{tabular}{l|p{4.cm}|c}
    \cmidrule[0.5pt]{1-3}
    \textbf{Edge Type} & \textbf{Description} & \textbf{Color Code}\\
    \cmidrule[0.5pt]{1-3}
    Structural Dependency & From an instruction mnemonic node to the instruction mnemonic node of the following instruction. &\tikz{\draw[dashed,line width=1.5pt, arrows={-Latex[length=3mm]}] (0,0)--(1,0);}\\\cmidrule[0.5pt]{1-3}
    Input Operand  & From a value node to an instruction mnemonic node.&\tikz{\draw[color=colorcolover,line width=1.5pt, arrows={-Latex[length=3mm]}] (0,0)--(1,0);}\\\cmidrule[0.5pt]{1-3}
    Output Operand & From an instruction mnemonic node to a register or a memory value node.& \tikz{\draw[color=colorblue,line width=1.5pt, arrows={-Latex[length=3mm]}] (0,0)--(1,0);}\\\cmidrule[0.5pt]{1-3}
    Address Base & From a register node to an address computation node. &\tikz{\draw[color=colorzereshk,line width=1.5pt, arrows={-Latex[length=3mm]}] (0,0)--(1,0);}\\\cmidrule[0.5pt]{1-3}
    Address Index & From a register node to an address computation node. &\tikz{\draw[dotted,color=colorzereshk,line width=1.5pt, arrows={-Latex[length=3mm]}] (0,0)--(1,0);}\\\cmidrule[0.5pt]{1-3}
    Address Segment & From a register node to an address computation node. &\tikz{\draw[dotted,color=colornaranj,line width=1.5pt, arrows={-Latex[length=3mm]}] (0,0)--(1,0);}\\\cmidrule[0.5pt]{1-3}
    Address Displacement & From an immediate value node to an address computation node.& \tikz{\draw[color=colornaranj,line width=1.5pt, arrows={-Latex[length=3mm]}] (0,0)--(1,0);}\\\cmidrule[0.5pt]{1-3}
    \end{tabular}
}
\end{table}
\begin{itempacked}
    \item \niparagraph{Node:} The initial feature vector of a node is a learnable embedding vector corresponding to the assembly language token associated with the node.
    The vector size of the node embedding vector is a model hyper-parameter.
    \item \niparagraph{Edge:}
    Similar to node initial embedding, the initial feature vector for an edge is a learnable embedding vector corresponding to the type of the edge.
    The vector size of the edge embedding vector is a model hyper-parameter.
    \item \niparagraph{Graph:}
    Finally, we also assemble an initial feature vector for the whole graph, called \emph{global feature} in \cite{graphnet:arxiv:2018}.
    The initial value of global feature vector indicates the relative frequencies of the tokens and edge types used in the graph. The size of global feature vector is equal to the number of token and edge types in the model.
\end{itempacked}

\subsection{Decoder Network}
\label{sec:decoder}
Once the graph embedding vectors are produced by the GNN, the decoder network uses these vectors to predict the estimated basic block throughput values.
The decoder is a multi-layer feed forward ReLU network with residual connections and layer normalization~\cite{ba2016layer} at input that is applied to the feature vector of each instruction mnemonic node and returns a scalar output.
Intuitively, an instruction mnemonic node represents the instruction, and the decoder network computes the contribution of the instruction to the overall throughput.
We sum the outputs of the decoder for each instruction mnemonic node to compute the throughput estimation for the whole basic block.

\subsection{Multi-task Decoder Network}
\label{sec:multi-task-decoder}
The multi-task version of the model uses the same decoder network architecture as the single-task decoder, but there is a separate decoder network for each target microarchitecture.
The graph neural network is shared across all tasks, learning a shared representation of basic blocks regardless of their target microarchitectures, whereas the dedicated decoder networks are used to predict the throughput values for different microarchitectures. 
\begin{figure}
    \centering
    \includegraphics[width=0.49\textwidth]{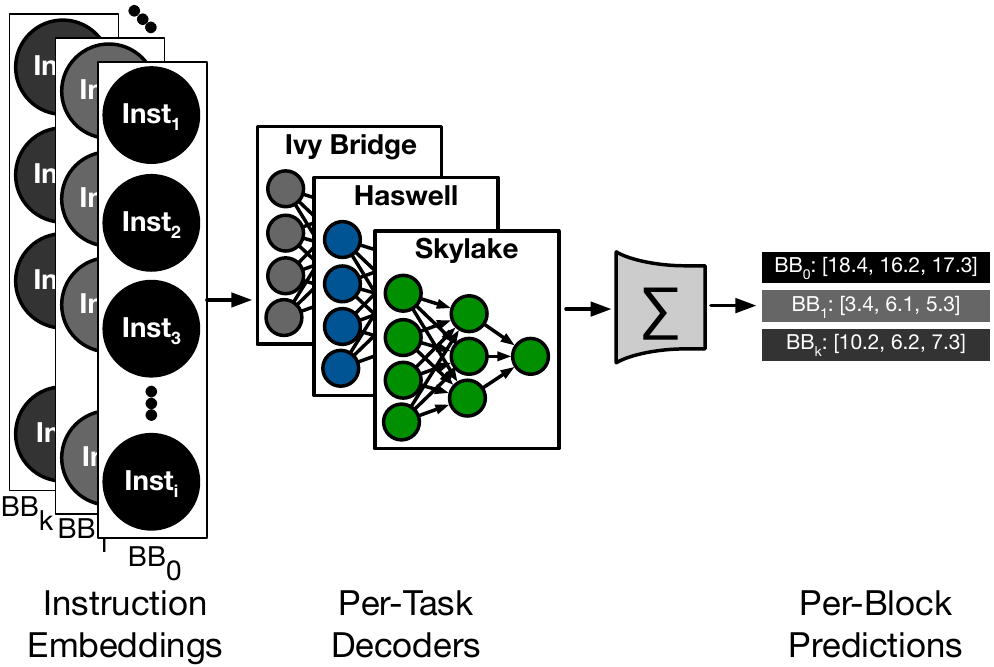}
    \caption{High-level schematic of the \oracle model with multi-task heads.}
    \label{fig:oracle_model}
\end{figure}
Figure~\ref{fig:oracle_model} shows a high-level architecture of a multi-task \oracle model.
\section{Methodology}
\label{sec:methodology}
\niparagraph{Dataset.}
We trained and tested the \oracle model on two existing datasets:
(1) the dataset used in the \ithemal paper~\cite{ithemal:icml:2019} with more than 1.4M basic blocks\footnote{The authors of the \ithemal paper~\cite{ithemal:icml:2019} kindly shared the dataset with us.} and 
(2) \bhive~\cite{bhive:iiswc:2019}, an open-source benchmark suite with more than 300K basic blocks.
Both datasets provide throughputs from measurements on three recent Intel microarchitectures: Ivy Bridge, Haswell, and Skylake.
However, \ithemal~\cite{ithemal:icml:2019} and \bhive~\cite{bhive:iiswc:2019} datasets were constructed using different measurement tools and it is challenging to blend the measurements from these datasets.
These datasets embody various domains, including database, compiler~\cite{lattner2004llvm} and performance optimization benchmarks~\cite{spec:2006,spec:2017}, scientific computing, and machine learning.

To evaluate, we randomly split each dataset into a training part comprising $83\%$ of blocks and a test part containing $17\%$ of blocks.
We use the \emph{same} split of the data set in all experiments to isolate the impact of dataset distribution on the final results.
When training the models, we let the training algorithm run for $\geq$6M training steps (roughly one week of real time).
We further split the training data into training ($98\%$) and validation ($2\%$). We use the validation split to select the best checkpoint during training.

\niparagraph{Implementation.}
We implemented \oracle using TensorFlow~1.x~\cite{tf} and DeepMind's Graph Nets library~\cite{graph_nets_library:web}.
For the embedding update functions, we used a two-layer feedforward ReLU network.
For the purpose of evaluation, we re-implemented the \ithemal~\cite{ithemal:icml:2019} model using the same version of TensorFlow\footnote{The source code of our implementations can be found under open-source license at \url{https://github.com/google/gematria}.}.
To ensure consistency between the models, we employ the same setup for training and evaluation.
In all the comparisons, we regard \ithemal~\cite{ithemal:icml:2019} as the baseline model.
We employ Mean Absolute Percentage Error (MAPE) as the loss function, an identical loss function to \ithemal~\cite{ithemal:icml:2019}:
\begin{equation*}
    \textrm{$\mathcal{L}$(actual, predicted)} = \frac{\lvert \textrm{actual - predicted} \rvert}{|\textrm{actual}|}
\end{equation*}
\noindent where \bench{actual} and \bench{predicted} indicate the measured throughput from hardware and predicted throughput from learned models, respectively.
We use Adam~\cite{adam:arxiv:2014} optimizer with a learning rate of 1e-3 and the default decay rates for moment estimations.
Table~\ref{tab:hparams} summarizes the rest of default hyperparameters and the architecture of learned models.
Unless otherwise specified, we use the default hyperparameter values in all the experiments.
\begin{table}[t!]
\small
\renewcommand{\arraystretch}{1.1}
\centering
\caption{\oracle hyper-parameters used during training.}
\label{tab:hparams}
\resizebox{0.49\textwidth}{!}{
\begin{tabular}{l|l}
      \textbf{Hyper-parameter} & \textbf{Value}\\
      \cmidrule[1.1pt]{1-2}
      Learning Rate & 1e-3\\
      Number of Basic Blocks / Batch & 100\\
      \hline
      Node Update Layers & 2$\times$256\\
      Node Embedding Size & 256\\
      \hline
      Edge Update Layers & 2$\times$256\\
      Edge Embedding Size & 256\\
      \hline
      Global Update Layers & 2$\times$256\\
      Global Embedding Size & 256\\
      \hline
      Task Decoder Layers & 2$\times$256\\
      \hline
      Number of Message Passing Iterations & 4-8\\
      Layer/Decoder Normalization & True\\
      Layer/Decoder Residual Connections & True\\
      \hline
      Aggregation Type & $\sum$ Node Embeddings\\
      \cmidrule[1.1pt]{1-2}
    \end{tabular}
}
\end{table}

\niparagraph{Extensions to the \ithemal model.}
The \ithemal model as described in \cite{ithemal:icml:2019} is trained to predict throughput values for a single microarchitecture.
In addition, the \ithemal~\cite{ithemal:icml:2019} model uses a single dot-product operation as its decoder network.
In our evaluations, we find that a multi-task decoder network using a multi-layer ReLU feed forward network can boost model accuracy.
To have a head-to-head comparison and isolate the impact of the GNN on the quality of the predictions, we augmented the \ithemal~\cite{ithemal:icml:2019} model with these extensions.
We add these extensions, replacing the single dot-product operation with the same decoder network as described in Section~\ref{sec:multi-task-decoder}.
We refer to this extended \ithemal model as \qk{\ithemalp}.
\section{Evaluation}
\label{sec:eval}
\subsection{Baseline Comparisons}
This section provides the model accuracy comparison results with baseline learned model~\cite{ithemal:icml:2019} on the \ithemal dataset.
In summary, \oracle outperforms \ithemal by a margin of roughly 1.7$\%$, and by 1.93$\%$ on average across all microarchitectures.

\niparagraph{Comparison with \ithemal.}
We evaluate the accuracy of \oracle, \ithemal (baseline), and \ithemalp{} (\ithemal with our proposed extensions) with respect to the ground truth throughput data across three x86-64 microarchitectures.
We trained all models on the \ithemal training dataset and we report their accuracy on \ithemal testing dataset.
Table~\ref{table:acc_comparison_ithemal} presents the accuracy comparisons and two correlation metrics (e.g. Spearman and Pearson).
The Spearman correlation metric measures the rank correlation between two variables, whereas Pearson correlation metric gauges the linear relation between them.
\begin{table}[t!]
\small
\renewcommand{\arraystretch}{1.1}
\centering
\caption{Comparison of best accuracy results achieved with \oracle (ours), \ithemal~\cite{ithemal:icml:2019}, and \ithemalp~when trained and tested on the \ithemal dataset. Bold and \underline{underline} values show the best and second best results, respectively.}
\label{table:acc_comparison_ithemal}
\resizebox{0.49\textwidth}{!}{
\begin{tabular}{l|l|l|l|l}
\cmidrule[1.5pt]{1-5}
\textbf{Dataset}&\textbf{Microarchitecture}&\textbf{Model}&\textbf{MAPE}& \textbf{Spearman / Pearson} \\
\cmidrule[1.5pt]{1-5}
\multirow{9}{*}{\ithemal}
&\multirow{3}{*}{Ivy Bridge}
&\ithemal & $8.34 \%$ & $0.9640$ / $0.2768$ \\
\cline{3-5}
&&\ithemalp& \underline{7.89}$\% $ & $ 0.9744$ / \textbf{0.9631} \\
\cline{3-5}
&&\oracle & \textbf{6.67}$\%$ & $ 0.9721$ / \underline{0.8936} \\
\cmidrule[1.5pt]{2-5}
&\multirow{3}{*}{Haswell}
& \ithemal\hspace{-1mm}$^*$ & $9.90 \%$ & $0.9720$ / $0.3615 $\\\cline{3-5}
&&\ithemalp& \underline{8.82}$\%$ & $0.9777$ / \textbf{0.9231}\\
\cline{3-5}
&&\oracle& \textbf{7.61}$\%$ & $ 0.9752$ / \underline{0.8255}\\\cmidrule[1.5pt]{2-5}
&\multirow{3}{*}{Skylake}&\ithemal& $8.30 \%$ & $ 0.9643$ / $0.2871 $ \\\cline{3-5}
&&\ithemalp & \underline{7.51}$\%$ & $ 0.9754$ / \textbf{0.9035}\\
\cline{3-5}
&&\oracle& \textbf{6.47}$\%$ & $ 0.9717$ / \underline{0.7888}\\
\cmidrule[1.5pt]{1-5}
\end{tabular}
}
\raggedright{\scriptsize{$^*$ We obtained the results by training the models with mean squared percentage error. The accuracy results when training the models with mean \emph{absolute} percentage error was significantly worse.}}
\end{table}
On the \ithemal testing dataset, \oracle outperforms \ithemal across all the microarchitectures by at least $1.67\%$, and by \xx{1.93$\%$} on average.

When model trained on the \ithemal dataset is tested on the \bhive dataset, the prediction accuracy for both learned models drops significantly.
This trend is expected because the \bhive dataset uses a different methodology to measure the throughput values.
Nevertheless, under this setting \oracle still yields lower prediction error in comparison to \ithemal model, on average, by \xx{0.39}$\%$.
The prediction accuracy on the \bhive dataset between \ithemalp{} and \oracle are comparable.
\oracle consistently outperforms \ithemalp{} on Ivy Bridge (\xx{10.47$\%$} vs. \xx{11.01$\%$}) and Skylake (\xx{11.26$\%$} vs. \xx{11.39$\%$}) microarchitectures, whereas \ithemalp{} yields marginally lower accuracy on Haswell microarchitecture (\xx{11.64$\%$} vs. \xx{11.57$\%$}).
All the learned models yield comparable Spearman correlations (\xx{0.96-0.98}).
However, we obtain the best Pearson correlations with \ithemalp{} and \oracle, significantly outperforming \ithemal vanilla model.

Table~\ref{table:acc_comparison_bhive} summarizes the test error when \oracle and \ithemalp are trained and tested on the \bhive dataset (with a proper split between and training and testing). We did not include vanilla \ithemal in this comparison because of consistent numerical instability in the training process.
\oracle{} consistently outperforms \ithemalp{} across the three microarchitectures in terms of test error as well as Pearson correlation.
On average, \oracle yields 0.64$\%$ lower test error, while providing considerably better Pearson correlation.
Both models yield comparable Spearman correlation.
\begin{table}[t!]
\small
\renewcommand{\arraystretch}{1.1}
\centering
\caption{Performance of \oracle, trained and tested on the \bhive dataset. The correlation metrics use the same terminology as \cite{ithemal:icml:2019}.}
\label{table:acc_comparison_bhive}
\resizebox{0.49\textwidth}{!}{
\begin{tabular}{l|l|l|l}
\cmidrule[1.pt]{1-4}
\textbf{Microarchitecture}&\textbf{Model}&\textbf{MAPE}& \textbf{Spearman / Pearson} \\
\cmidrule[1.pt]{1-4}
\multirow{2}{*}{Ivy Bridge}
&\ithemalp& $9.25\%$ & $ 0.9725$ / $0.5424$\\\cmidrule[0.5pt]{2-4}
&\oracle& \textbf{8.44}$\%$ & $ 0.9593$ / \textbf{0.9873}\\\cmidrule[1.5pt]{1-4}
\multirow{2}{*}{Haswell}
&\ithemalp& $9.19\%$ & $ 0.9684$ / $0.5334 $\\\cmidrule[0.5pt]{2-4}
&\oracle& \textbf{8.41}$\%$ & $ 0.9550$ / \textbf{0.9633}\\\cmidrule[1.5pt]{1-4}
\multirow{2}{*}{Skylake}
&\ithemalp& $9.45\%$ & $0.9698$ / $0.8402$\\\cmidrule[0.5pt]{2-4}
&\oracle& \textbf{9.12}$\%$ & $ 0.9524$ / \textbf{0.9423}\\\cmidrule[1.pt]{1-4}
\end{tabular}
}
\end{table}

\begin{figure*}[t]
\centering
\subfloat[Ivy Bridge - \ithemal]{\label{fig:ivy:ithemal}\includegraphics[width=0.33\textwidth]{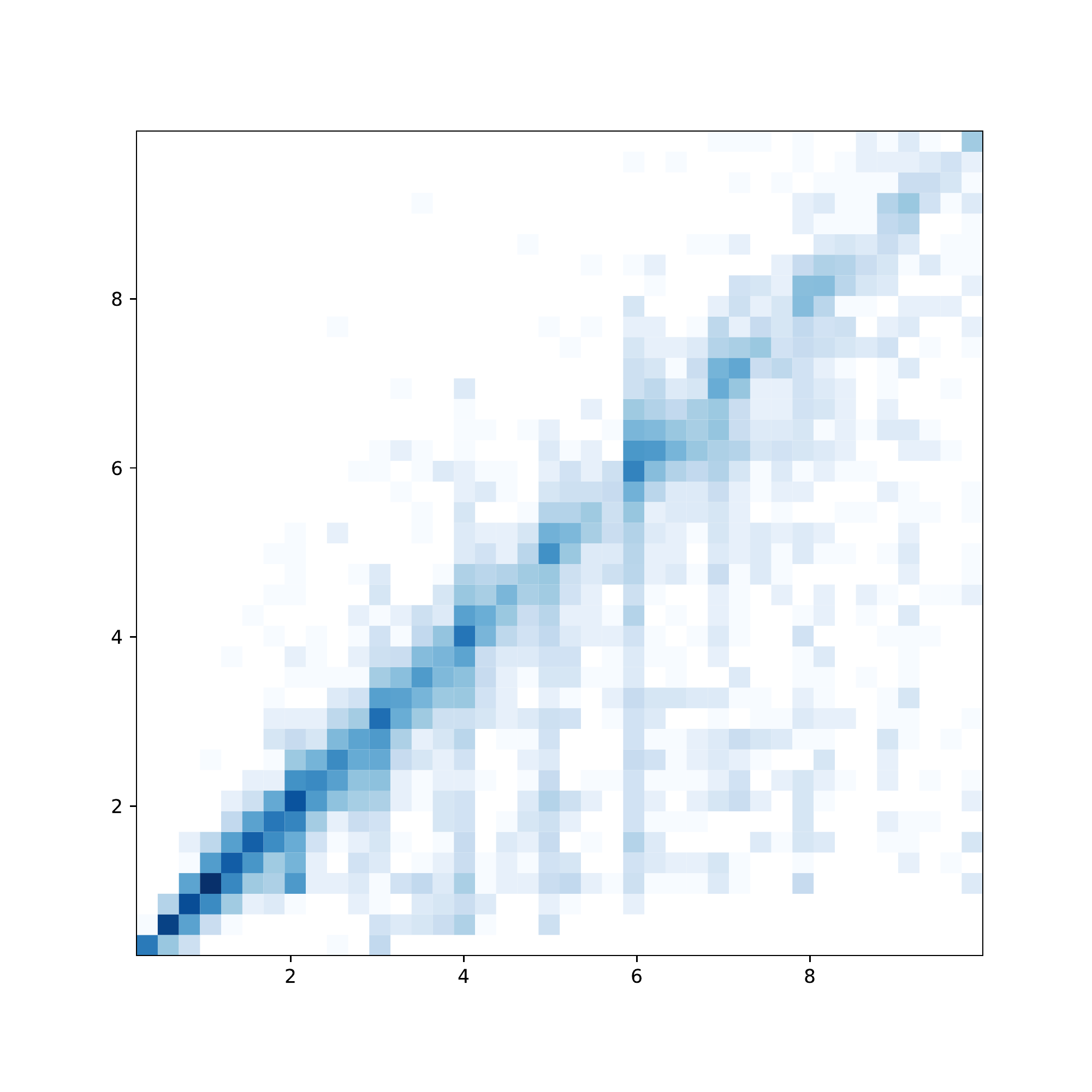}}
\subfloat[Haswell - \ithemal]{\label{fig:has:ithemal}\includegraphics[width=0.33\textwidth]{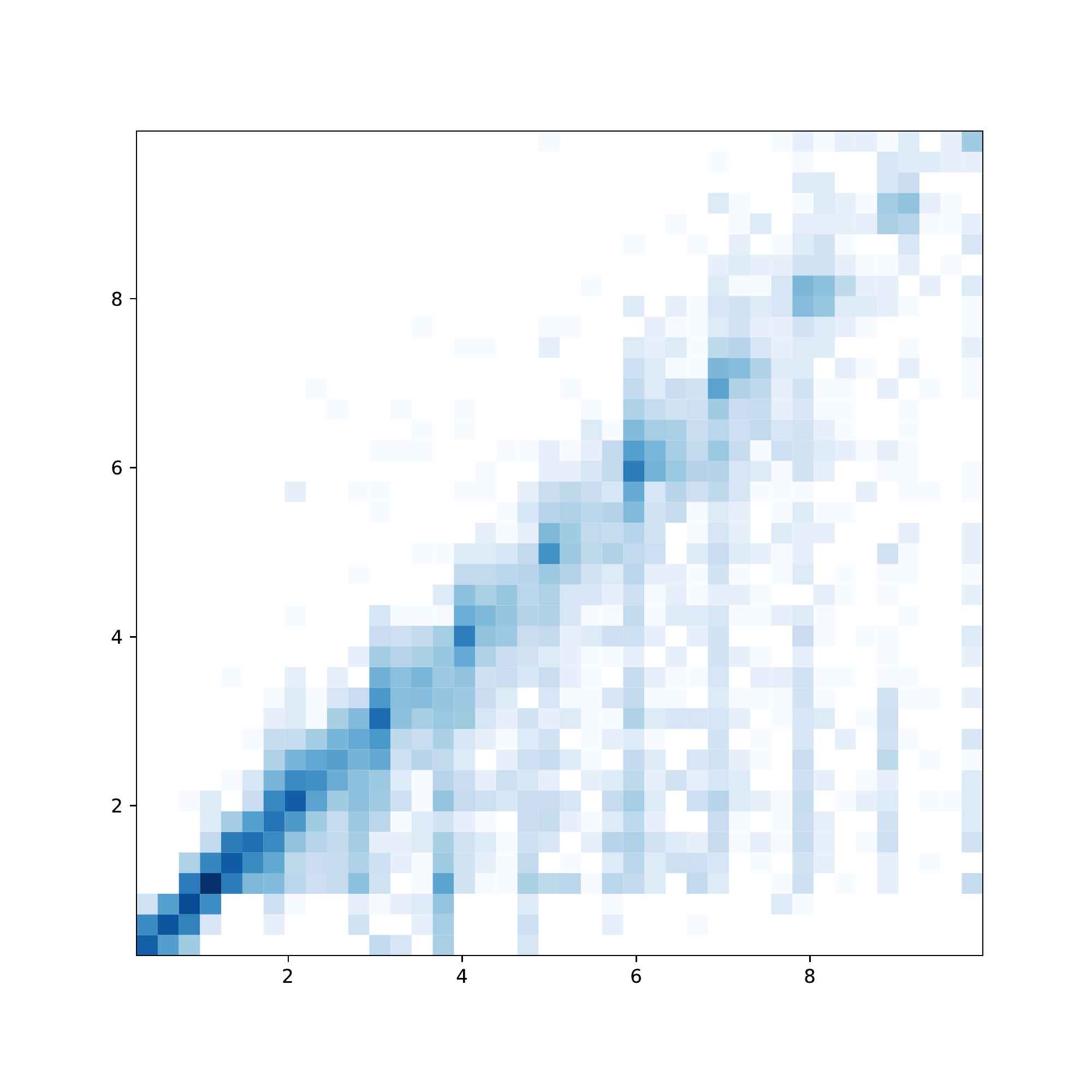}}
\subfloat[Skylake - \ithemal]{\label{fig:sky:ithemal}\includegraphics[width=0.33\textwidth]{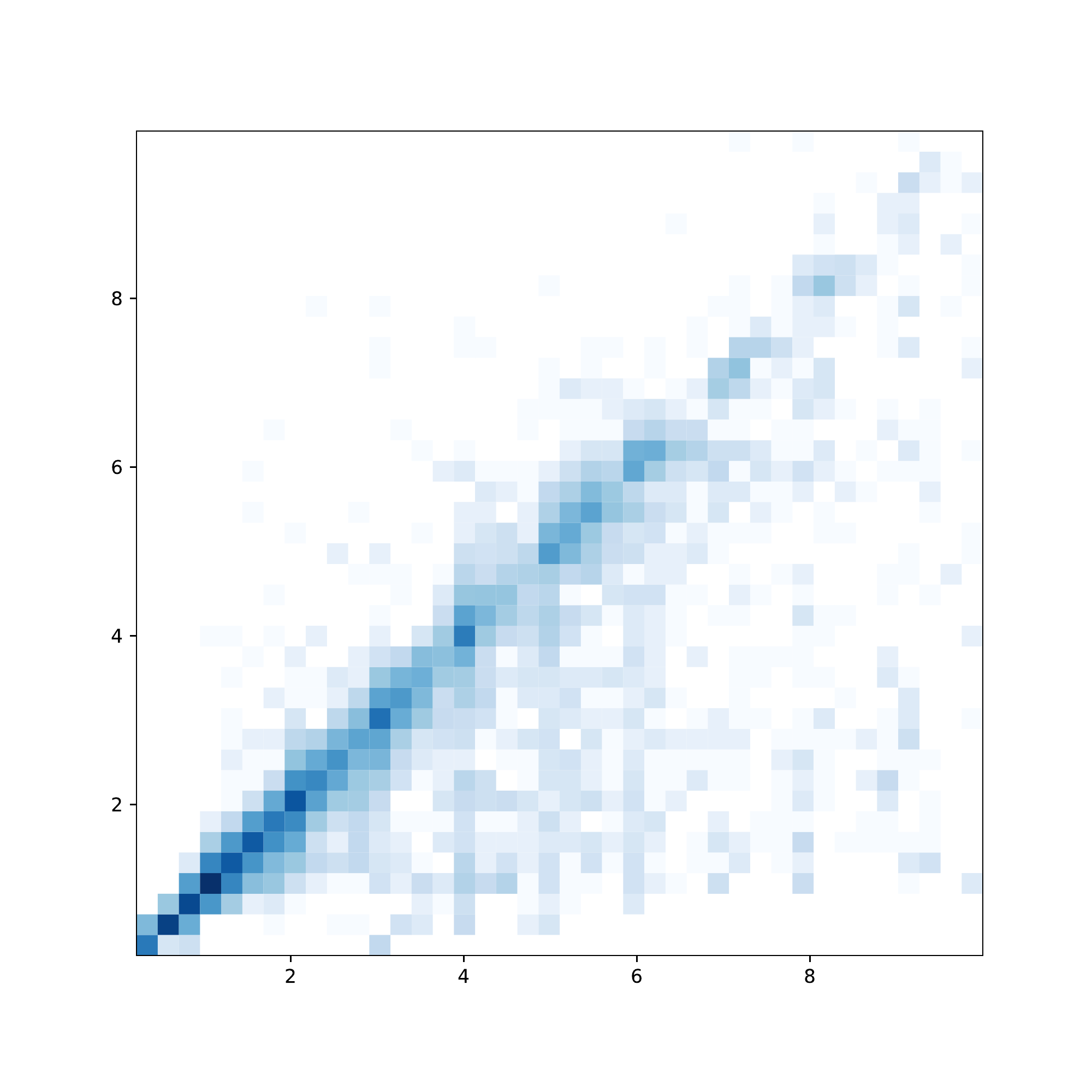}}\\
\subfloat[Ivy Bridge - \oracle]{\label{fig:ivy:oracle}\includegraphics[width=0.33\textwidth]{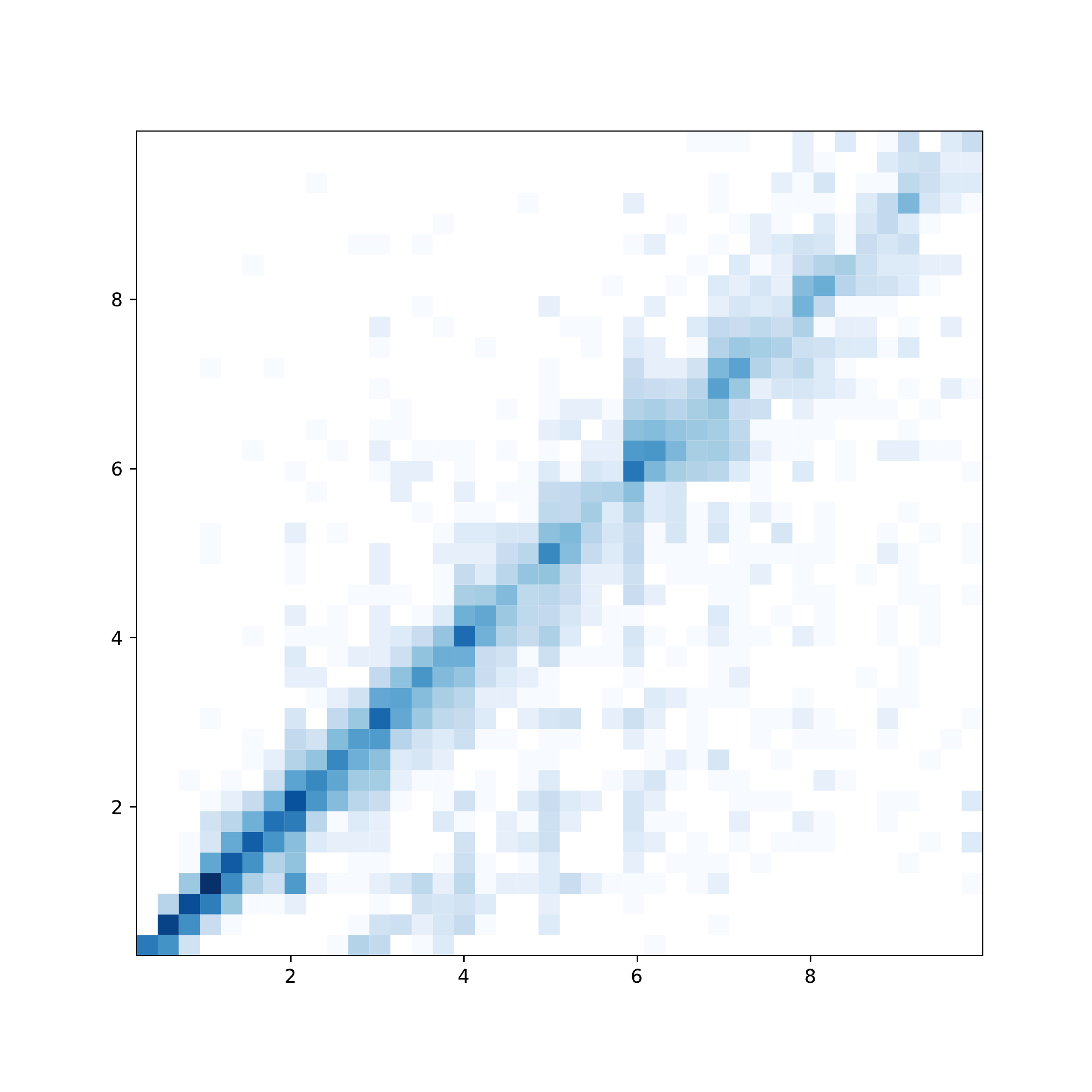}}
\subfloat[Haswell - \oracle]{\label{fig:has:oracle}\includegraphics[width=0.33\textwidth]{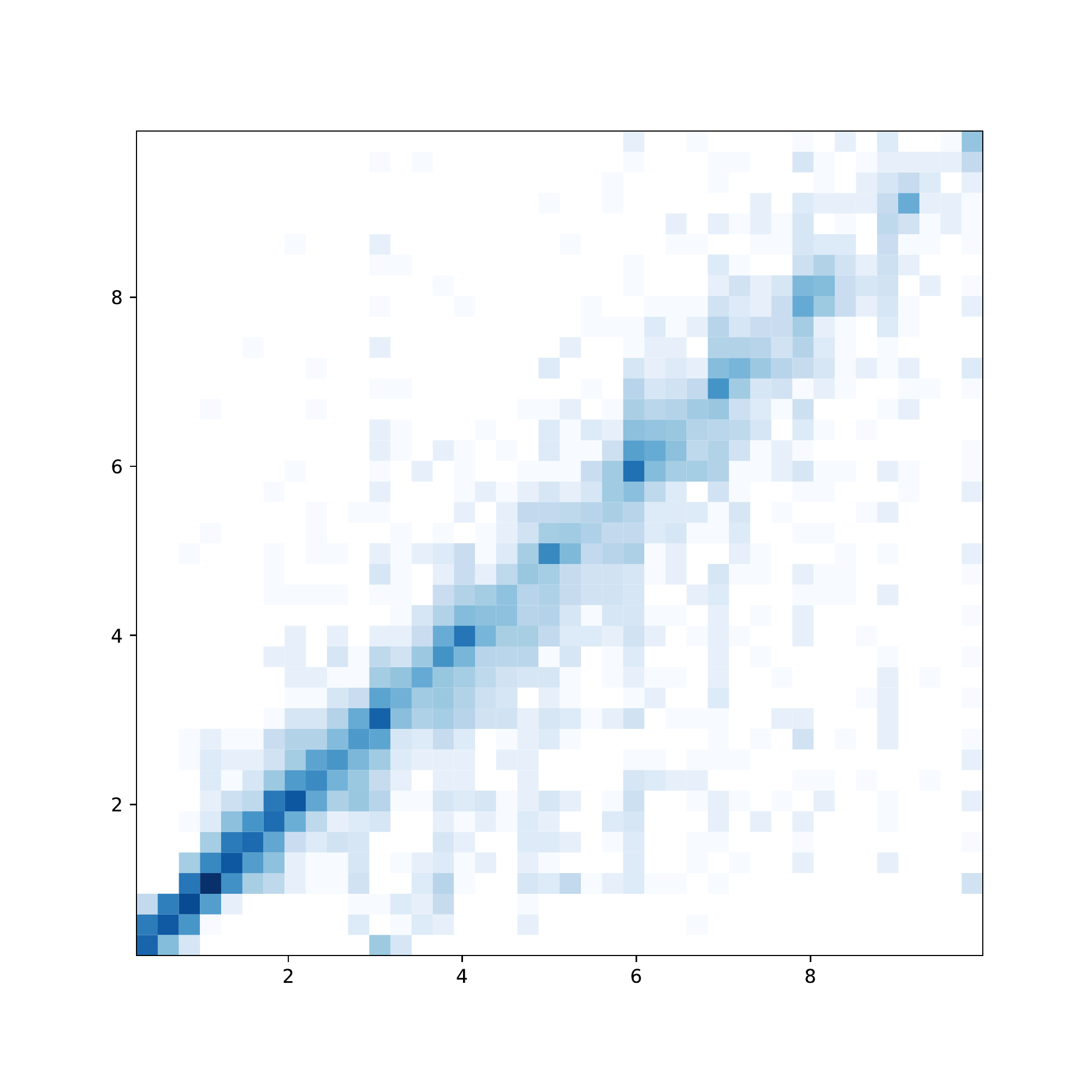}}
\subfloat[Skylake - \oracle]{\label{fig:sky:oracle}\includegraphics[width=0.33\textwidth]{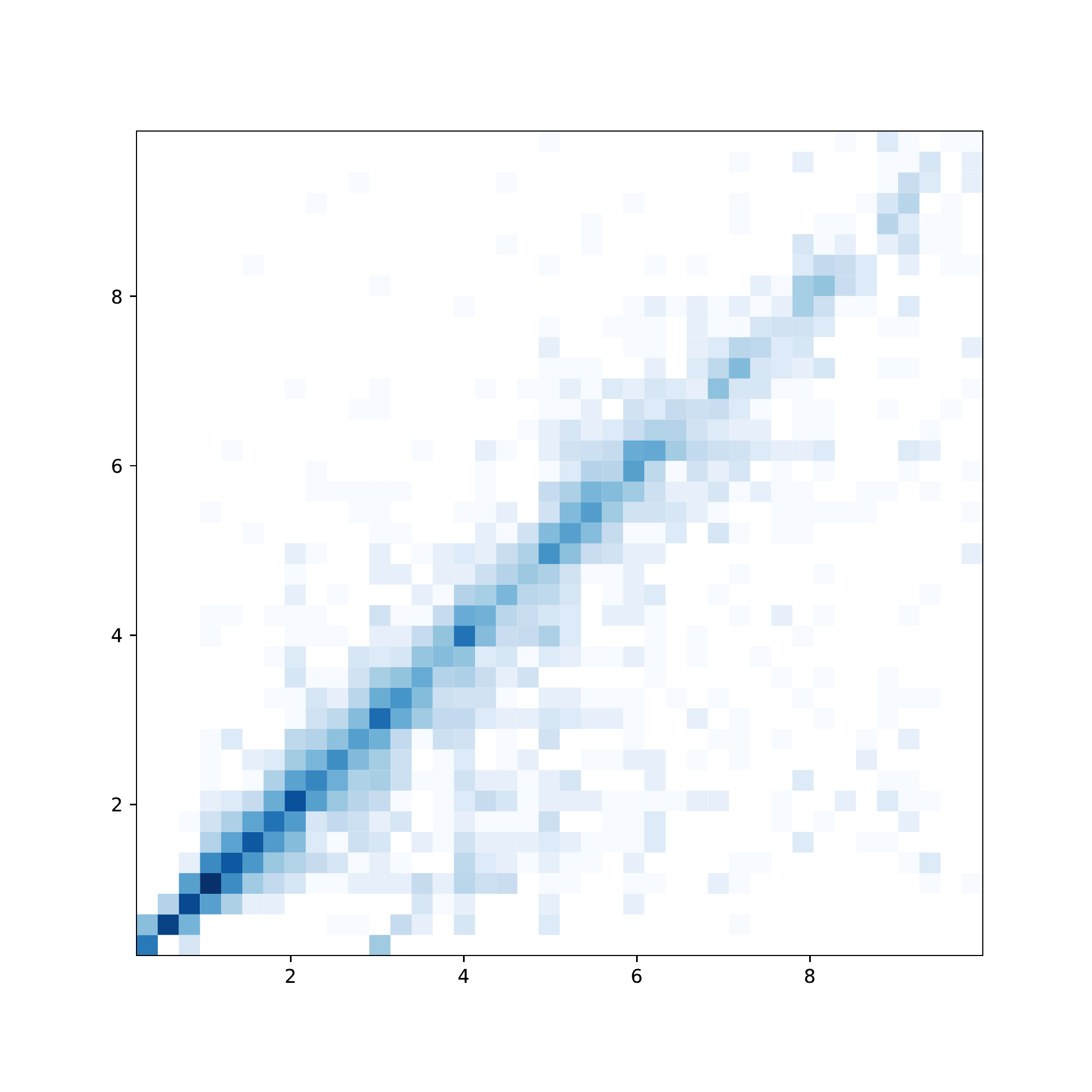}}\\
\caption{Heatmaps for ground-truth (x axis) and predicted values (y axis) for \ithemal~\cite{ithemal:icml:2019} and multi-task \oracle learned models on the \ithemal dataset~\cite{ithemal:icml:2019} across three different x86-64 microarchitectures for the throughput values under 10 cycles.}
\label{fig:heatmap:ithemal_ds} 
\end{figure*}

\begin{figure*}[t]
\centering
\subfloat[Ivy Bridge - \ithemal]{\label{fig:ivy:ithemal_error_dist}\includegraphics[width=0.32\textwidth]{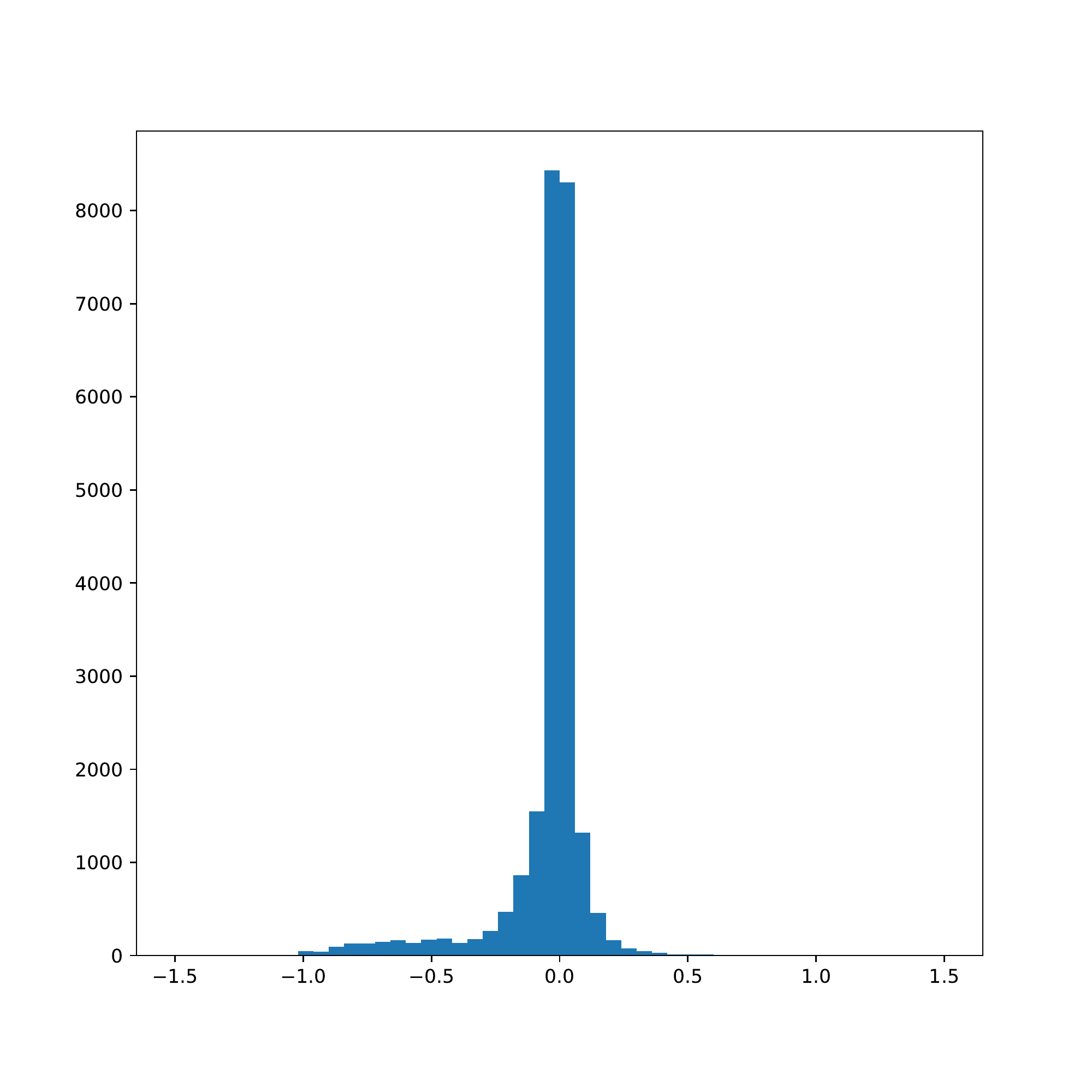}}
\subfloat[Haswell - \ithemal]{\label{fig:has:ithemal_error_dist}\includegraphics[width=0.32\textwidth]{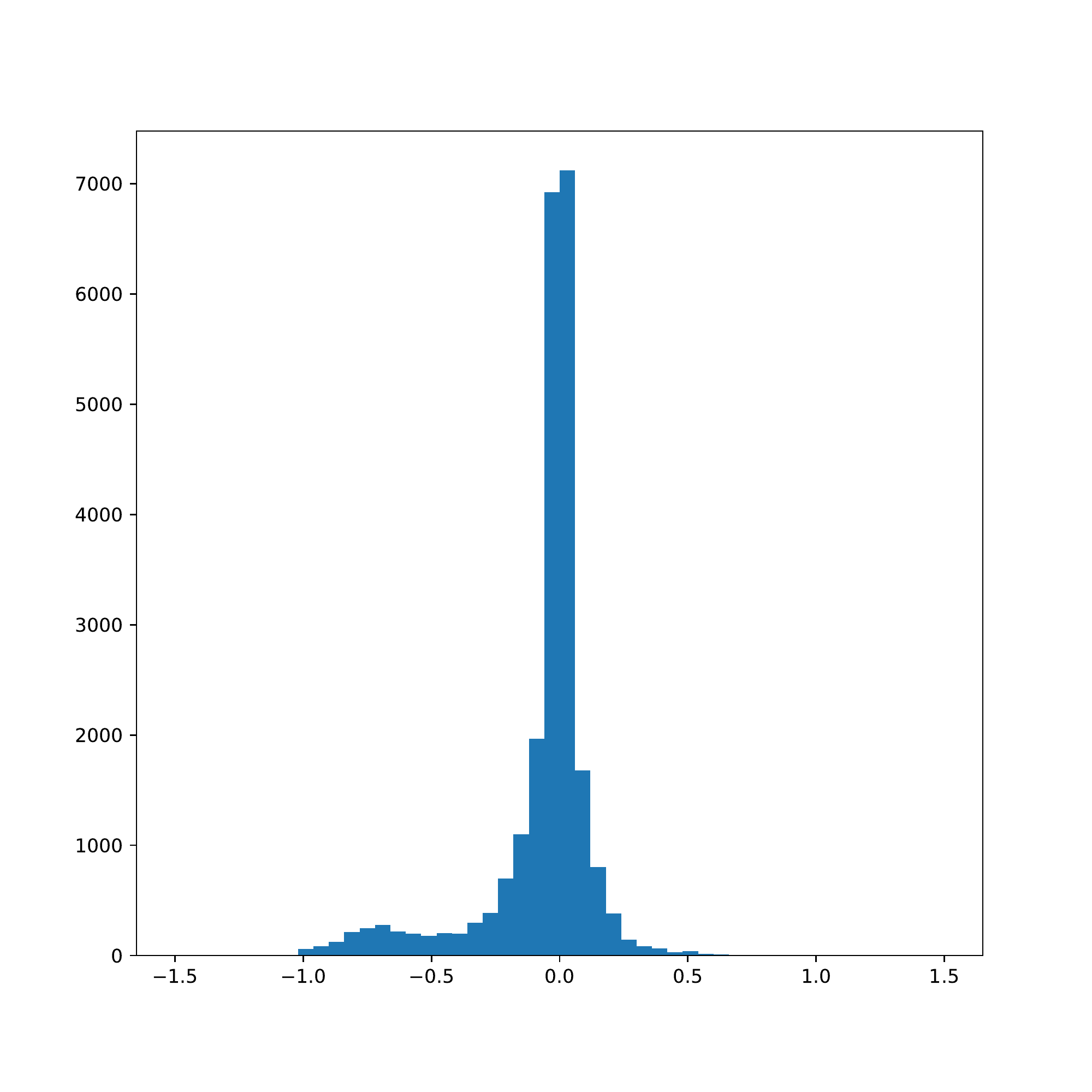}}
\subfloat[Skylake - \ithemal]{\label{fig:sky:ithemal_error_dist}\includegraphics[width=0.32\textwidth]{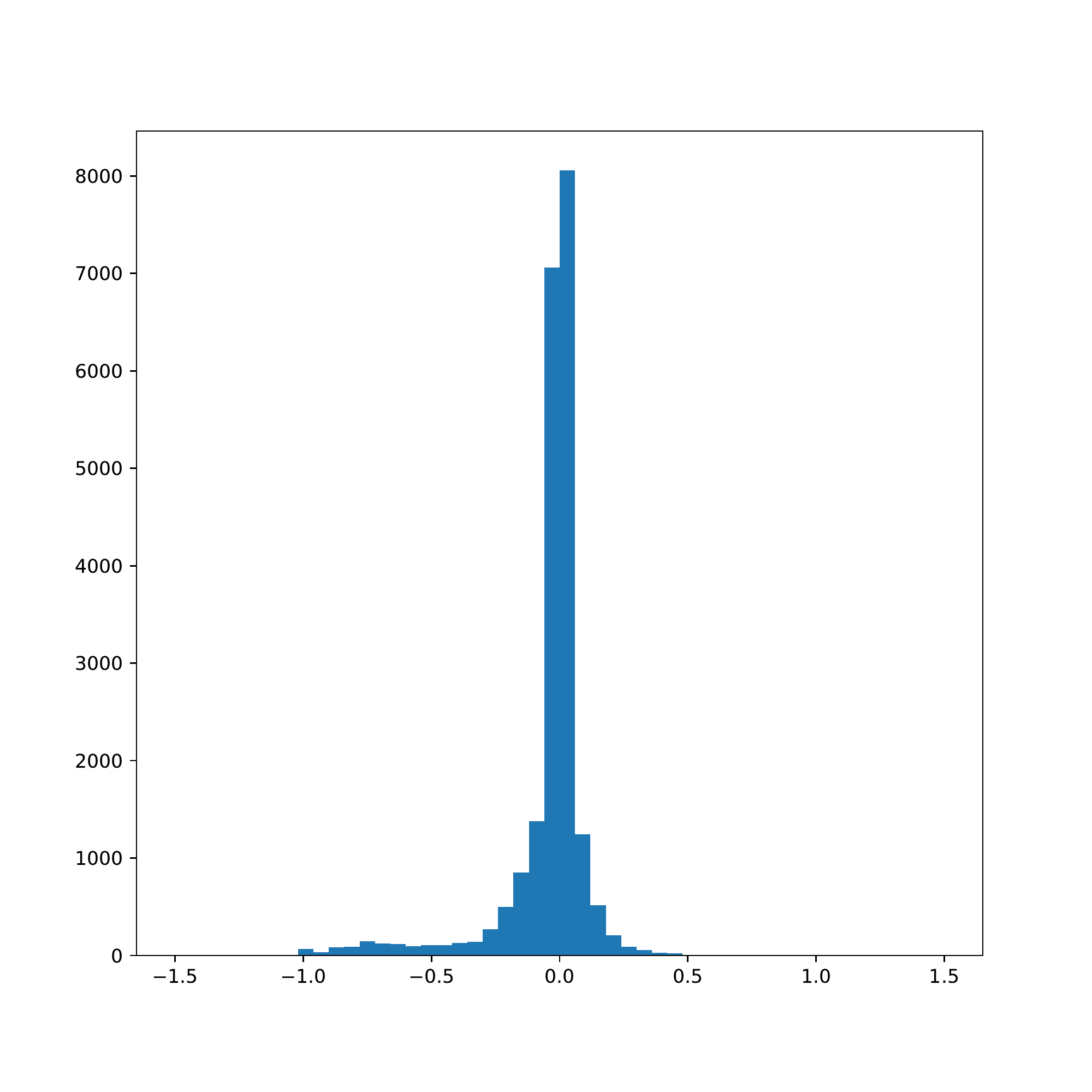}}\\
\subfloat[Ivy Bridge - \oracle]{\label{fig:ivy:oracle_error_dist}\includegraphics[width=0.32\textwidth]{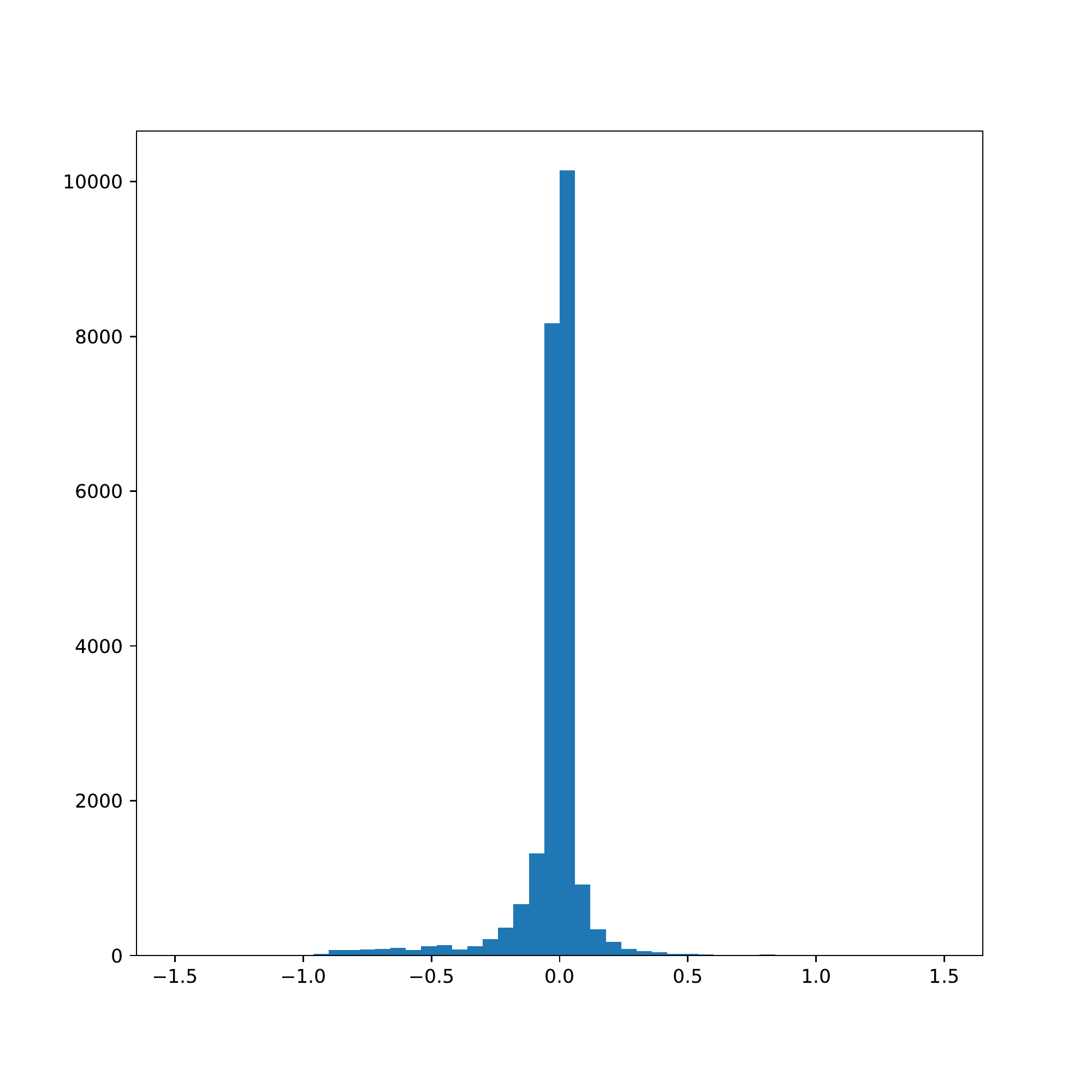}}
\subfloat[Haswell - \oracle]{\label{fig:has:oracle_error_dist}\includegraphics[width=0.32\textwidth]{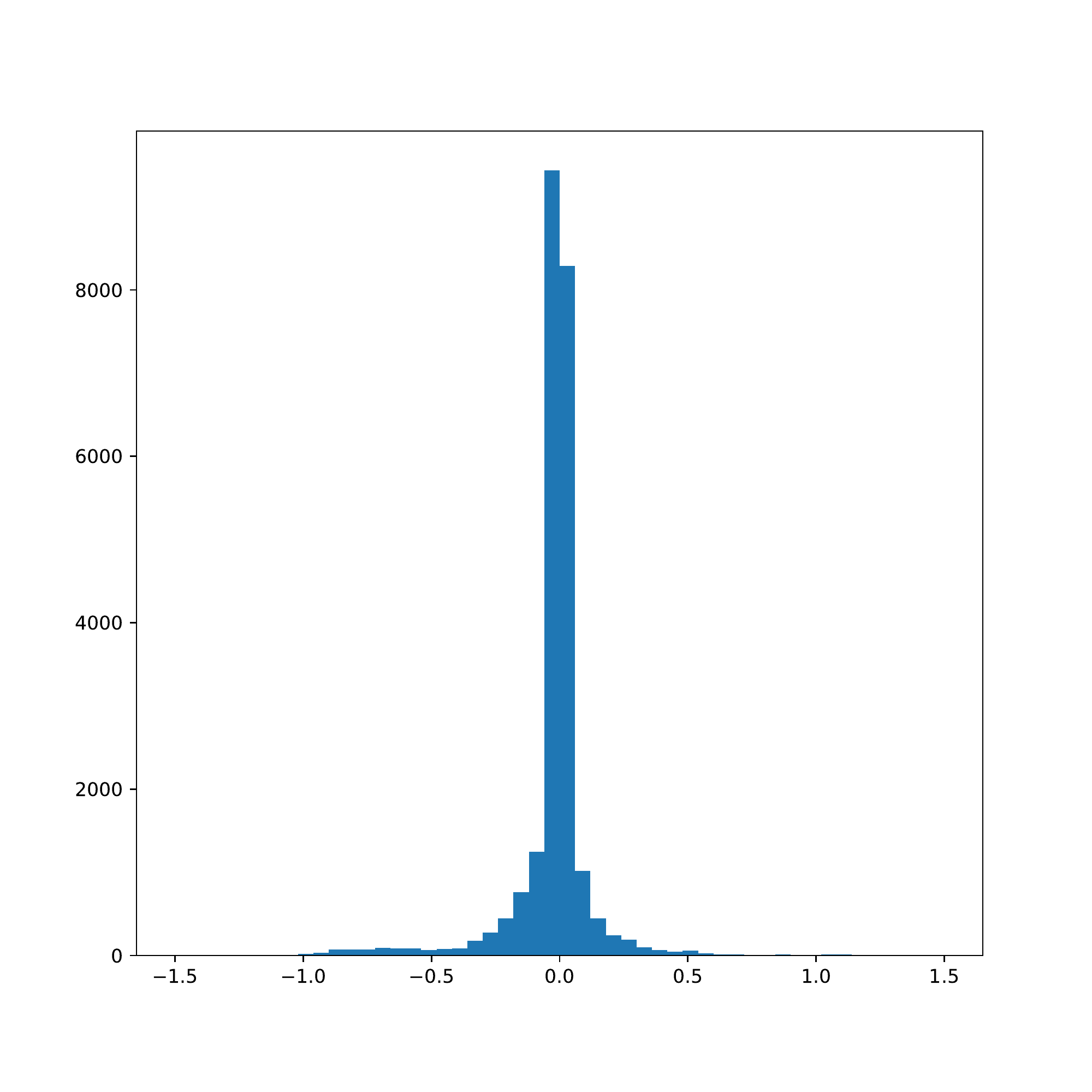}}
\subfloat[Skylake - \oracle]{\label{fig:sky:oracle_error_dist}\includegraphics[width=0.32\textwidth]{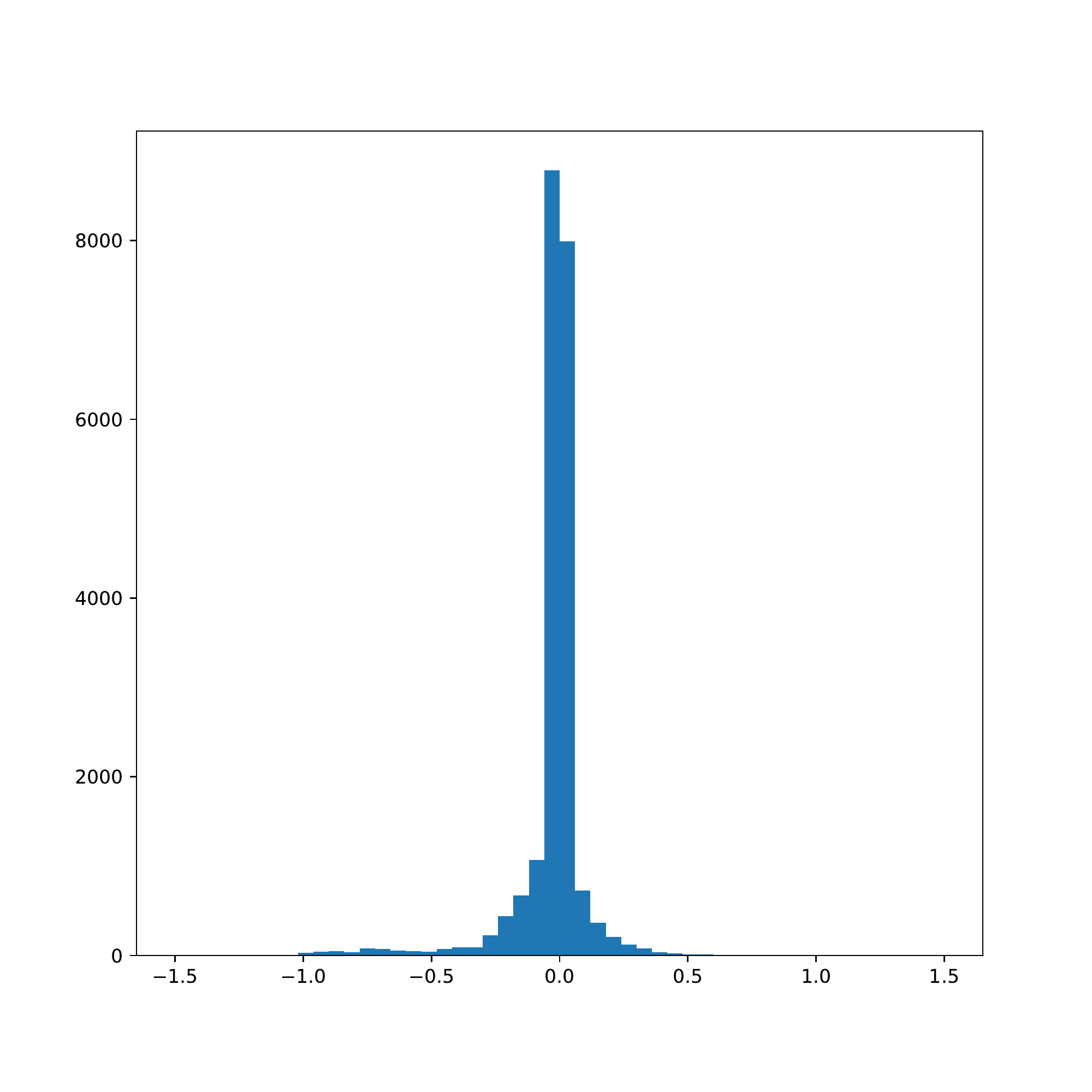}}\\
\caption{Distribution of relative error (x axis) and the number of basic blocks (y axis) for \ithemal~\cite{ithemal:icml:2019} and multi-task \oracle learned models on the \ithemal dataset~\cite{ithemal:icml:2019} across three different x86-64 microarchitectures, corresponding to heatmaps in Figure~\ref{fig:heatmap:ithemal_ds}.}
\label{fig:heatmap:ithemal_ds_error_dist} 
\end{figure*}

\niparagraph{Analysis of learned models on \ithemal dataset.}
Figure~\ref{fig:heatmap:ithemal_ds} shows the prediction heatmap analysis for each microarchitecture.
We use the same methodology as \cite{ithemal:icml:2019} to obtain the heatmaps, except we normalize the throughput values to a single run of each basic block.
The first row shows the results for \ithemal, whereas the second one shows the results for \oracle.
Our model uniformly yields higher density along the $y$ = $x$ line (perfect estimator line).
The \ithemal model has a tendency to underestimate (higher density under the $y$ = $x$ line), which is avoided by \oracle.
We conjecture that this is due to the per-instruction decoding of the \oracle model.
To better illustrate this behavior, Figure~\ref{fig:heatmap:ithemal_ds_error_dist} shows the distribution of relative errors of both models across various microarchitectures.


\begin{figure*}[t]
\centering
\subfloat[Ivy Bridge - \oracle]{\label{fig:ivy:oracle_bhive}\includegraphics[width=0.33\textwidth]{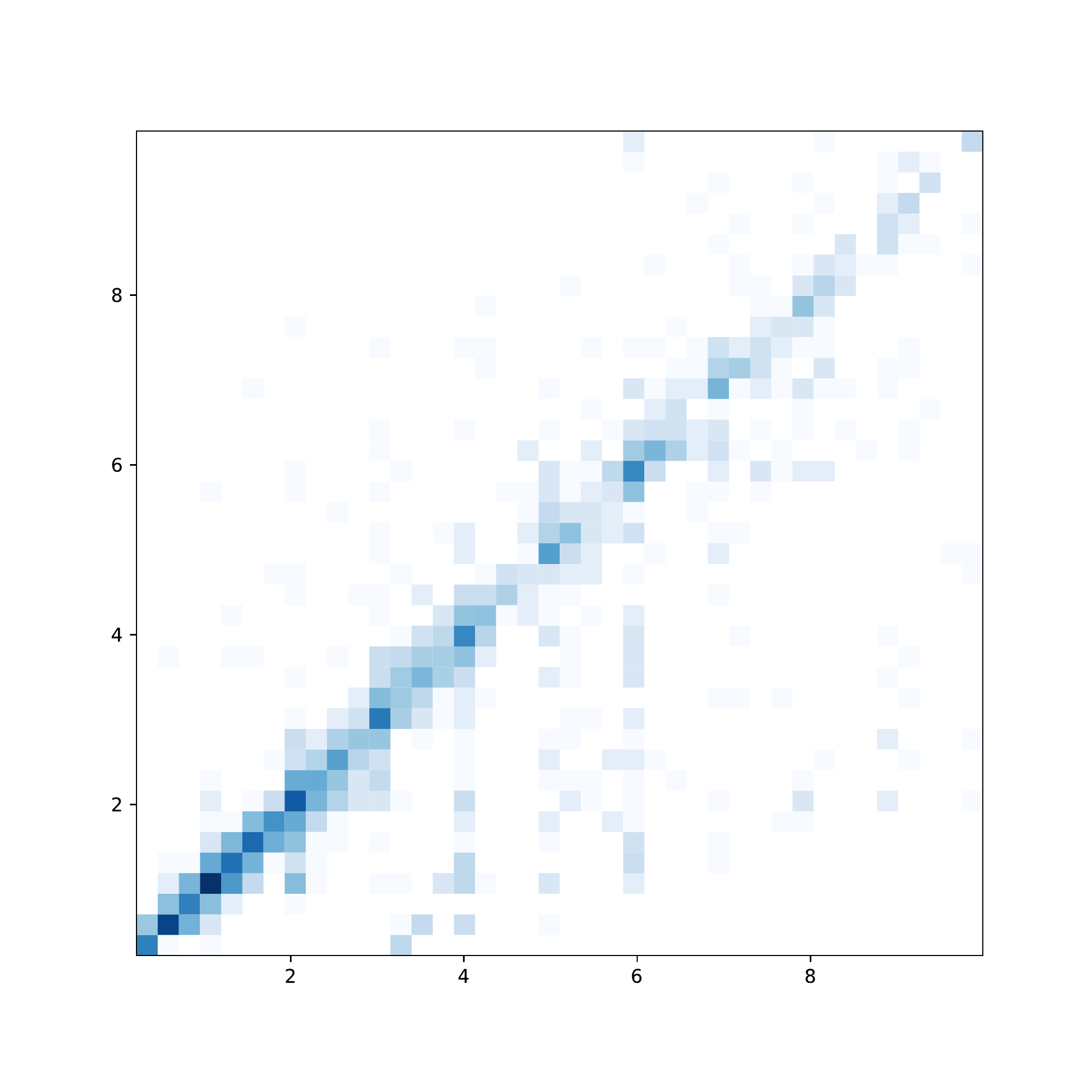}}
\subfloat[Haswell - \oracle]{\label{fig:has:oracle_bhive}\includegraphics[width=0.33\textwidth]{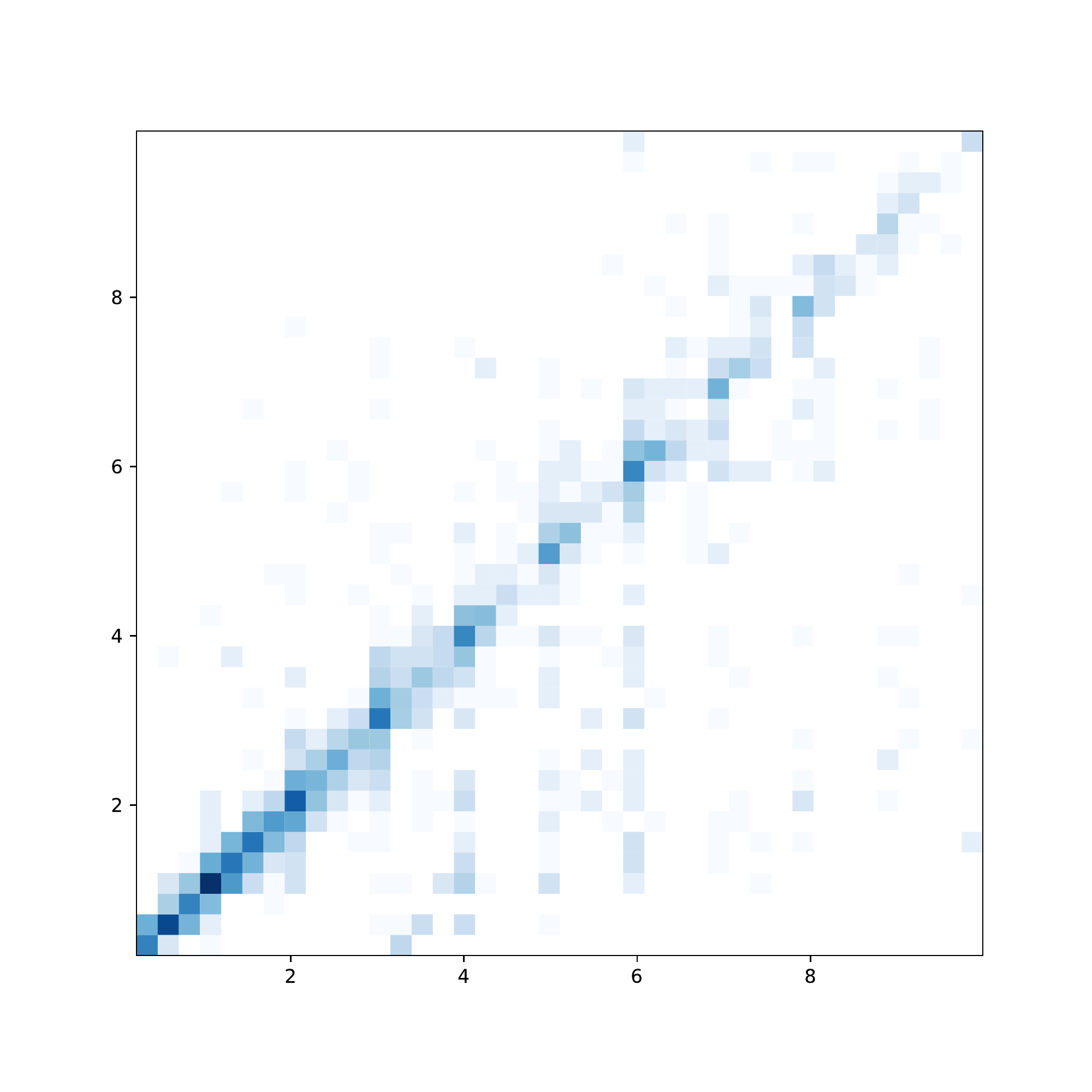}}
\subfloat[Skylake - \oracle]{\label{fig:sky:oracle_bhive}\includegraphics[width=0.33\textwidth]{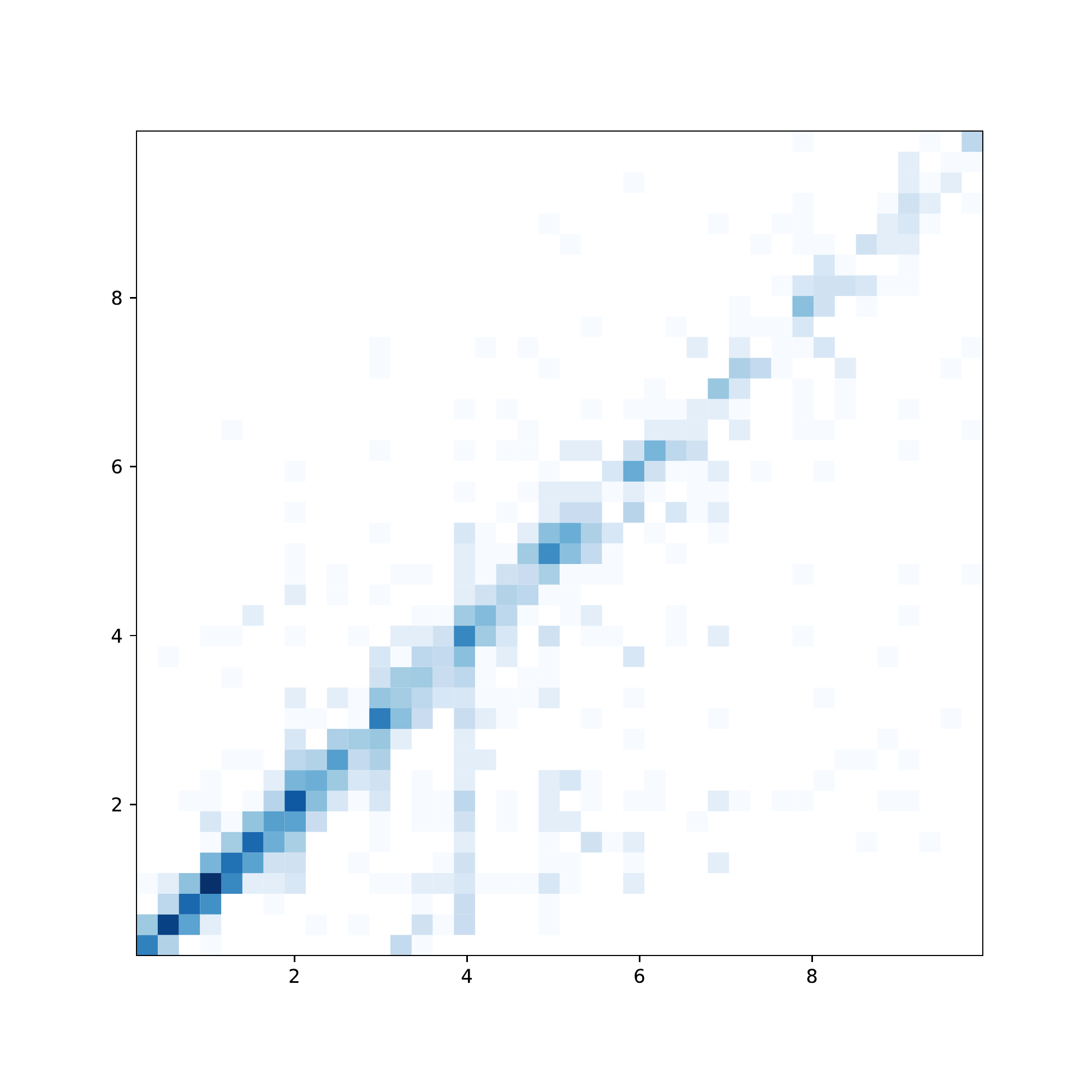}}\\
\caption{Heatmaps for ground-truth (x axis) and predicted values (y axis) for \ithemal and \oracle learned models on the \bhive dataset across three different x86-64 microarchitectures for the throughput values under 10 cycles. Note that the \bhive~\cite{bhive:iiswc:2019} dataset contains 5$\times$ less data compared to the \ithemal dataset~\cite{ithemal:icml:2019}; Hence the sparser heatmaps.}
\label{fig:heatmap:bhive_ds} 
\end{figure*}

\niparagraph{Analysis of learned models on the \bhive dataset.}
Figure~\ref{fig:heatmap:bhive_ds} illustrates the same analysis for \oracle model when trained and tested on the \bhive dataset.
Note that the \ithemal data set is 5$\times$ bigger than the \bhive dataset; hence, the heatmaps in Figure~\ref{fig:heatmap:bhive_ds} appear to be sparser than heatmaps in Figure~\ref{fig:heatmap:ithemal_ds}.
Similar to the trend observed in Figure~\ref{fig:heatmap:ithemal_ds}, \oracle on the \bhive dataset yields a comparable performance between underestimated and overestimated predicted values.
These detailed analysis of the learned models indicates that \oracle consistently outperforms \ithemal~\cite{ithemal:icml:2019} across the measured throughput spectrum.
\begin{table}[t!]
\small
\renewcommand{\arraystretch}{1.1}
\centering
\caption{Sensitivity of \oracle to the number of message passing iterations on \ithemal dataset~\cite{ithemal:icml:2019} across three different microarchitectures.}
\label{table:num_message_passings}
\resizebox{0.47\textwidth}{!}{
\begin{tabular}{l|c|l}
\cmidrule[1.1pt]{1-3}
\textbf{Microarchitecture}&\textbf{\# of Message Passing Iterations}&\textbf{Mean Absolute Error}\\\cmidrule[1.1pt]{1-3}
\multirow{3}{*}{Ivy Bridge}&1& $8.48 \%$\\\cline{2-3}
&2& $7.85 \%$\\\cline{2-3}
&4& $7.49 \%$\\\cline{2-3}
&8& \textbf{6.67}$\%$\\\cline{2-3}
&12& $7.30 \%$\\\cmidrule[1.1pt]{1-3}
\multirow{3}{*}{Haswell}&1& $9.42 \%$\\\cline{2-3}
&2& $9.09 \%$ \\\cline{2-3}
&4& $8.40 \%$\\\cline{2-3}
&8& \textbf{7.61}$\%$\\\cline{2-3}
&12& $8.44 \%$\\\cmidrule[1.1pt]{1-3}
\multirow{3}{*}{Skylake}&1& $8.40 \%$ \\\cline{2-3}
&2& $7.47 \%$ \\\cline{2-3}
&4& $7.05 \%$ \\\cline{2-3}
&8& \textbf{6.47}$\%$ \\\cline{2-3}
&12& $6.97 \%$ \\\cmidrule[1.1pt]{1-3}
\end{tabular}
}
\end{table}
\subsection{Ablation Studies}
In this section, we perform detailed ablation studies across various hyper-parameters of the learned model and summarize our observations.

\niparagraph{Sensitivity to the number of message passing iterations.}
We first sweep the number of message passing iterations in the graph neural network with one, two, four, eight, and twelve.
Each message passing iteration constitutes a synchronous exchange of embedding vectors between adjacent graph nodes and edges.
The number of message passing iterations limits the distance that information from each node and edge can ``travel'' in the graph.
Increasing the number of iterations allows information exchange between more distant nodes, but at the same time it makes training and inference more computationally expensive.

Table~\ref{table:num_message_passings} summarizes the results across all three microarchitectures.
With eight message passing iterations, \oracle achieves the lowest test prediction error, on average, $6.67\%$ and $0.65\%$ lower than \oracle with two and six message passing iterations, respectively.
The results show that \oracle's performance is indeed sensitive to the number of message passing iterations, suggesting a search to find the sweet spot for this hyper-parameter.
We postulate that as the number of nodes (instructions) increases, a higher number of message passing iterations could potentially further reduce the prediction error, owing to better capturing the underlying dependencies between nodes.
However, increasing the number of message passing iterations more than a certain value (eight in our setup) could lead to a higher inductive bias to training dataset.

\niparagraph{Impact of the decoder network.}
To determine the effect of the decoder network on the quality of the model, we 
modified the \ithemal model~\cite{ithemal:icml:2019} to use the same multi-layer feed forward network with ReLU non-linearity.
We observed that adding the decoder network improved the \ithemalp model accuracy by $0.25 \%$, $0.39 \%$, and $1.1 \%$ for Ivy Bridge, Haswell, and Skylake, respectively.
We attribute these improvement to the additional non-linearity of the decoder network that incorporate more inductive bias to the network.
We observe a similar trend in multi-task training, where multi-task learning is most effective in the presence of multi-layer feed forward network.
The following provides a possible explanation for this trend.
In the vanilla \ithemal model~\cite{ithemal:icml:2019}, a simple dot-product operation is used to model the throughput computation.
We hypothesize that such modeling imposes the task of throughput prediction as well as semantic modeling of basic block dependencies onto LSTM layers, which possibly hinders the capability of the model to construct an expressive representation of basic blocks.

\niparagraph{Sensitivity to layer normalization.}
Layer normalization has proved to be effective in stabilizing the training for recurrent neural networks~\cite{ba2016layer}.  
As part of the sensitivity study, we explore the impact of layer normalization on \oracle accuracy.
For this experiment, we remove all the layer normalization from node and edge update networks and the decoder network.
The results show that without layer normalization the test prediction error, significantly increases by $15.19 \%$, $12.87 \%$, and $12.27\%$ for Ivy Bridge, Haswell, and Skylake microarchitectures, respectively.
We also observed that disabling layer normalization significantly increases numerical instability that we had to counter by using gradient clipping.
This significant increase in the test prediction error suggests the importance of layer normalization in achieving the state-of-the-art accuracy for basic block throughout estimation as well as improving the numerical stability of the training.

\niparagraph{Sensitivity to loss function.}
Finally, we analyze the impact of various loss functions on the final model error.
The vanilla \ithemal~\cite{ithemal:icml:2019} model trains and evaluates the models using MAPE.
However, employing different loss functions may potentially lead to better generalization to unseen data and less overfitting~\cite{li2018visualizing}. 
To verify that MAPE is indeed the best loss function for \oracle, we trained the model with other loss functions and evaluated the model MAPE against an equivalent model trained with MAPE.
The additional loss functions that we studied are mean squared error (MSE) and Huber loss~\cite{huber} in two setups: (1) with absolute error, calculated as the difference between predicted and ground-truth values and (2) relative error, computed as the absolute error normalized by the ground-truth value.
Compared to MSE, the Huber loss is known to be less sensitive to outliers in the dataset. In all the experiments with Huber loss, we set $\delta$ = 1.

Table~\ref{table:loss_comparison} summarizes the comparison between different loss function across different microarchitectures.
We report various comparison metrics (columns three to six of Table~\ref{table:loss_comparison}) for each loss function.
While training with MAPE generally provides best results, we observe that relative MSE may also be a viable option.
Other loss functions and in particular loss functions that do not use normalization perform significantly worse due to the high dynamic range of the predicted throughput values.

\subsection{Multi-Task Learning}
\begin{table}[t!]
\small
\renewcommand{\arraystretch}{1.1}
\centering
\caption{The effects of multi-task training on \oracle and \ithemal models~\cite{ithemal:icml:2019} across different x86-64 microarchitectures.}
\label{table:multitask}
\resizebox{0.49\textwidth}{!}{
\begin{tabular}{l|l|l|l}
\cmidrule[1.1pt]{1-4}
\textbf{Microarchitecture}&\textbf{Model}&\textbf{MAPE (Single-Task)}&\textbf{MAPE (Multi-Task)}\\\cmidrule[1.1pt]{1-4}
\multirow{3}{*}{Ivy Bridge}&\ithemal& $8.34 \%$ & $8.82 \%$ \\\cline{2-4}
&\ithemalp& $8.37 \%$ & $7.89 \%$ \\\cline{2-4}
&\oracle& \underline{7.02}$\%$ & \textbf{6.67}$\%$ \\\cmidrule[1.1pt]{1-4}
\multirow{3}{*}{Haswell}&\ithemal& $9.90 \%$ & $9.62 \%$\\\cline{2-4}
&\ithemalp & $8.87 \%$ & $8.82 \%$ \\\cline{2-4}
&\oracle & \textbf{7.76} $\%$ & \underline{7.82}$\%$\\\cmidrule[1.1pt]{1-4}
\multirow{3}{*}{Skylake}&\ithemal& $8.30 \%$ & $8.77 \%$ \\\cline{2-4}
&\ithemalp& $7.65 \%$ & $7.51 \%$ \\\cline{2-4}
&\oracle& \underline{7.34} $\%$ & \textbf{6.75} $\%$ \\\cmidrule[1.1pt]{1-4}
\end{tabular}
}
\end{table}
In this section, we evaluate the impact of multi-task learning on performance prediction.
Each task in this context represents a target microarchitecture (e.g. Ivy Bridge, Haswell, and Skylake).
Our goal is to explore whether it is feasible to design a generalized model that works across different microarchitectures, likely with disparate characteristics. 
When training a multi-task model, we selected basic blocks where we had ground truth data for all target microarchitectures.
In addition, for each basic block, we update the weights for all target microarchitectures at the same time.

Table~\ref{table:multitask} compares the performance of \oracle multi-task model and \ithemalp with multi-task heads.
It shows that in most cases using multi-task learning (1) improves the quality of the trained model, and (2) it makes training more efficient by training a single model for multiple microarchitectures at once.
The main case where multi-task learning has negative impact on the results is in case of the unmodified \ithemal model~\cite{ithemal:icml:2019}.
We attribute this to the simplicity of the task-specific decoder part in this model; we see that when the model is augmented with a more complex task-specific decoder, the model can benefit from multi-task training.
We also take this as an indication that the shared part of the network learns a representation of code that is sufficiently powerful to support multiple target microarchitectures.

\begin{table*}[t]
\small
\renewcommand{\arraystretch}{1.1}
\centering
\caption{Comparison between different loss functions in \oracle on \ithemal dataset~\cite{ithemal:icml:2019}. Note that in our data sets, throughput values are per 100 iterations of each basic block which explains higher MSE and Huber loss values.}
\label{table:loss_comparison}
\begin{tabular}{l|l|r|r|r|r|r}
\cmidrule[1.1pt]{1-7}
\textbf{Microarchitecture}& \textbf{Loss Function} & \textbf{MAPE} & \textbf{MSE} & \textbf{Relative MSE} & \textbf{Mean Huber} & \textbf{Mean Relative Huber}\\\cmidrule[1.1pt]{1-7}
\multirow{5}{*}{Ivy Bridge}& MAPE & \textbf{7.49}$\%$ & 2353023.37 & 0.926 & 91.23 & 0.022 \\ \cline{2-7}
& MSE & $24.94 \%$ & 1709602.44 & 1.670 & 124.28 & 0.072 \\\cline{2-7}
&Relative MSE & \underline{7.72}$\%$ & 1922472.84 & \textbf{0.044} & \textbf{86.85} & \textbf{0.016} \\\cline{2-7}
&Huber& $10.21 \%$ & 1941646.88 & 0.966 & 87.52 & 0.036 \\\cline{2-7}
&Relative Huber& $8.34 \%$ & \textbf{1702852.03} & 0.676 & 88.72 & 0.022 \\\cmidrule[1.1pt]{1-7}
\multirow{5}{*}{Haswell}& MAPE & \textbf{8.33}$\%$ & 4883716.03 & 0.923 & 146.7 & 0.024 \\ \cline{2-7}
& MSE & $27.07 \% $ & 16328409.43 & 2.651 & 221.21 & 0.092 \\\cline{2-7}
&Relative MSE & \underline{8.88}$\%$ & 4138913.04 & \textbf{0.056} & 145.62 & \textbf{0.019} \\\cline{2-7}
&Huber&$11.51 \%$ & 4175191.19 & 0.931 & \textbf{142.59} & 0.039 \\ \cline{2-7}
&Relative Huber& $9.44 \%$ & \textbf{3777885.85} & 0.632 & 147.65 & 0.025 \\\cmidrule[1.1pt]{1-7}
\multirow{5}{*}{Skylake}& MAPE & \underline{7.32}$\%$ & 1407284.56 & 0.651 & 83.52 & 0.021 \\ \cline{2-7}
& MSE & $26.78 \%$ & 1202691.79 & 1.570 & 110.10 & 0.086 \\\cline{2-7}
&Relative MSE & \textbf{7.31}$\%$ & 1282483.60 & \textbf{0.032} & 80.24 & \textbf{0.013} \\\cline{2-7}
&Huber&$9.54 \%$ & \textbf{820971.73} & 0.579 & \textbf{66.44} & 0.029 \\ \cline{2-7}
&Relative Huber& $7.93 \%$ & 1334057.40 & 0.491 & 81.31 & 0.019 \\ \cmidrule[1.1pt]{1-7}
\end{tabular}
\end{table*}

\subsection{Computational efficiency}
Last, we consider the computational efficiency of the models. Efficiency is an important aspect of machine learning models deployed in practical applications.
We have evaluated the training and inference throughput of \oracle and compared it to the efficiency of other models discussed in this paper.
We used a Linux workstation with an Intel Xeon E5-1650-v3 CPU running at 3.50GHz, 128GB RAM, and an NVIDIA RTX 2080 Ti GPU.
For training, we report the average time per batch over 300 training steps of each model, whereas for inference, we report the average time per batch on the whole \bhive data set of ca 300k basic blocks. In both cases, we used batches of 100 basic blocks.

Table~\ref{table:runtimes} summarizes our results. Overall, we found that \oracle is roughly 3x faster than \ithemal{} and \ithemalp{} models both in training an inference when running on a GPU. When running inference on a CPU, \oracle is 27\% slower. We did not include CPU training time in our evaluation based on the observation that training is virtually always done using GPUs or other accelerators.

Moreover, our measurement have also shown that the overhead of training a multi-task models is negligible compared to training a similar model for a single task both for \oracle{} and models based on the \ithemal{} architecture. That is, the training cost per microarchitecture of a multi-task model with three heads is almost one third of the cost of training three equivalent single-task models.

\begin{table*}
\small
\centering
\caption{Run time per batch of 100 blocks of training and inference (in seconds)}
\label{table:runtimes}
\begin{tabular}{l|l|l|l|l}
\cmidrule[1.1pt]{1-5}
\textbf{Model} & \textbf{Microarchitecture} & \textbf{GPU training} & \textbf{GPU inference} & \textbf{CPU inference} \\
\cmidrule[1.1pt]{1-5}
\multirow{3}{*}{\ithemal single task} & Ivy Bridge & 0.0996s & 0.0491s & 0.0551s \\
 & Haswell & 0.1236s & 0.0501s & 0.0558s \\
 & Skylake & 0.0775s & 0.0502s & 0.0556s \\
\cmidrule[1.1pt]{1-5}
\multirow{3}{*}{\oracle single task} & Ivy Bridge & 0.0354s & 0.0147s & 0.0749s \\
 & Haswell & 0.0367s & 0.0147s & 0.0750s \\
 & Skylake & 0.0349s & 0.0146s & 0.0751s \\
\cmidrule[1.1pt]{1-5}
\ithemalp multi-task & Ivy Bridge \& Haswell \& Skylake & 0.1086s & 0.0515s & 0.0602s \\
\cmidrule[1.1pt]{1-5}
\oracle multi-task & Ivy Bridge \& Haswell \& Skylake & 0.0361s & 0.0157s & 0.0768s \\
\cmidrule[1.1pt]{1-5}
\end{tabular}
\end{table*}
\section{Related Work}
\label{sec:related}
\oracle takes a fundamentally different approach than the prior proposals for performance estimation of basic blocks.
In contrast to prior performance estimation work, \oracle takes one step further and leverages graph neural network theory to obtain expressive architecture embedding that translates to higher accuracy in the learned models.
Below, we overview the most relevant work.

\niparagraph{Performance estimation.}
There is a growing body of work on developing models for performance estimation that can be categorized into analytical models~\cite{critical_path:pmbs:2019,iaca:web:2019,llvm-mca:web:2019,throughput_intel:pmbs:2018,zsim:archnews:2013,marss:dac:2011,thread:hpca:2009,throughput_sc:tc:2008,chronos:scp:2007,perf_parallel:tocs:2004,pred_parallel:pc:2004,wcet:es:2001,pred_parallel:ipds:1998,static_perf:sc:1993,pred_static:rts:1993,bound:tools:1992} and learning based models~\cite{tpucost:arxiv:2020,halide:tog:2019,ithemal:icml:2019,quant_game:tecs:2012,gametime:tools:2011,regression:nips:2010,compiler_perf:cf:2007,edgetpu}.
Generally, developing analytical models is an intricate and tedious task in terms of human development, require meticulous understanding of internal microarchitectural details, and are rarely generalizable to different architectures.
In contrast, \oracle is a learning based model that aims to mitigate these challenges by leveraging machine learning techniques.

In the learned model category, \ithemal~\cite{ithemal:icml:2019} is the closest work to this paper in terms of overall approach. \ithemal uses a sequential LSTM-based model in which only the structural dependencies between adjacent instructions are present in an explicit form.
In contrast, \oracle uses GNNs to capture both short- and long-range dependencies between instructions in a graph representation of the basic block. In addition, this work takes one step further and, for the first time to the best of our knowledge, presents multi-task learning~\cite{caruana1997multitask} for throughput estimation across different architectures.
Kaufman et al.~\cite{tpucost:arxiv:2020} introduce a GNN-based performance model for tensor computation graphs on TPUs~\cite{tpu:isca:2017}. %
While tensor computation kernels are more complex than straight-line code like basic blocks, the in-order execution model of TPUs and the lack of hardware caching simplifies the task significantly. In contrast, \oracle targets throughout prediction in architectures with complex out-of-order execution models and multi-level caching.

\niparagraph{Graph neural networks.}
There is a growing interest of using graph neural networks in various reasoning tasks and to construct expressive low-dimensional representations from graph structures~\cite{graphnet:arxiv:2018,g2seq:arxiv:2018,gconvarg:aaai:2018,gnn_survey:arxiv:2018,semanticg:arxiv:2018,gconv:arxiv:2017,reason:nips:2017,crossg:tacl:2017,crossg:tacl:2017,rel-lstm:arxiv:2016,physics:nips:2016}. These learned low-dimensional representations are then generally processed to estimate desired metrics.
Computer programs can be represented naturally as graphs in which the nodes are associated with different elements of the assembly language representation of the code and the edges model different dependencies between these elements.
Recent work~\cite{code2vec:popl:2019,ncf:iclr:2019,graph-prog:arxiv:2017,Cummins2020ProGraMLGD} explores the idea of constructing graphs from source code and shows the strength of graph neural networks in various prediction tasks. 
As a natural step, we also use graphs to represent the dependencies in basic blocks and leverages the recent progress in graph neural networks~\cite{gnn_survey:arxiv:2018} to construct expressive representations for throughput estimation.

\section{Conclusion}
\label{sec:conclusion}
We present \oracle, a graph neural network model that establishes the state-of-the-art model accuracy for throughput estimation of basic blocks across various x86-64 microarchitectures.
Our results show that \oracle estimates the throughput of basic blocks with average test error of \xx{6.91}$\%$ across different microarchitectures, \xx{1.7}$\%$ over the previous state-of-the-art model~\cite{ithemal:icml:2019}, while also achieving ~3x higher throughput in training and inference, which can be further multiplied by training a single model to predict throughput for multiple target microarchitectures at the same time.
We have achieved these results by bringing in ideas from other fields of machine learning, such as graph neural networks~\cite{graphnet:arxiv:2018} and multi-task learning~\cite{caruana1997multitask}.

These promising results reinforce our claim about the expressiveness of the low-dimensional representations of basic blocks using graph neural networks.
We argue that using graphs to represent programs not only leads to richer low-dimensional representations which translate to higher accuracy and better generalization, but also paves the way to associate low-level microarchitectural features, such as performance counters, to each instruction.
This relational association between low-level microarchitectural features and programs is an exciting future research direction.
\section*{Acknowledgments}
\label{sec:ack}
We would like to extend our gratitude towards Jon Orwant, Corinna Cortes, Cliff Young, James Laudon, Stella Aslibekyan, the ``Learn to Design Accelerators'' team, the EXEgesis team, and the extended Google Research Brain Team for their invaluable feedback and comments.
\bibliographystyle{style/IEEEtran}
\bibliography{paper}
\end{document}